# Foxtsage vs. Adam: Revolution or Evolution in Optimization?

Sirwan A. Aula[1] and Tarik A. Rashid[2]

[1]Soran University, Computer Science Department, Soran, Erbil, Iraq
sirwan.aula@soran.edu.iq
[2]Computer Science & Engineering Department, Artificial Intelligence & Innovation Centre, University of Kurdistan Hewler, Erbil, Iraq
Corresponding author.
tarik.ahmed@ukh.edu.krd

## Abstract

Optimization techniques are pivotal in neural network training, shaping both predictive performance and convergence efficiency. This study introduces Foxtsage, a novel hybrid optimisation approach that integrates the Hybrid FOX-TSA with Stochastic Gradient Descent for training Multi-Layer Perceptron models. The proposed Foxtsage method is benchmarked against the widely adopted Adam optimizer across multiple standard datasets, focusing on key performance metrics such as training loss, accuracy, precision, recall, F1-score, and computational time. Experimental results demonstrate that Foxtsage achieves a 42.03% reduction in loss mean (Foxtsage: 9.508, Adam: 16.402) and a 42.19% improvement in loss standard deviation (Foxtsage: 20.86, Adam: 36.085), reflecting enhanced consistency and robustness. Modest improvements in accuracy mean (0.78%), precision mean (0.91%), recall mean (1.02%), and F1-score mean (0.89%) further underscore its predictive performance. However, these gains are accompanied by an increased computational cost, with a 330.87% rise in time mean (Foxtsage: 39.541 seconds, Adam: 9.177 seconds). By effectively combining the global search capabilities of FOX-TSA with the stability and adaptability of SGD, Foxtsage presents itself as a robust and viable alternative for neural network optimization tasks.

# 1. Introduction

## 1.1. Overview of Nature-Inspired Algorithms

NIAs have proven indispensable tools for solving complex optimisation challenges from engineering design to data science, to machine learning (Waleed et al. 2014). Inspired by such natural phenomena as biological evolution, animal behaviour and ecological systems. These algorithms use these natural patterns of search space exploration and use in order to explore and exploit such complex search spaces effectively (Campelo and Aranha 2023). Particle Swarm Optimization (PSO) (Kennedy, Eberhart, and gov 1995), Genetic Algorithms (GA) (Katoch, Chauhan, and Kumar 2021), and Grey Wolf Optimizer (GWO) are well-known NIAs and have been widely used for a wide number of problems. Nevertheless, there is no algorithm that is universally superior and hence a hybrid approach combining the merits of several algorithms is required (Hauben et al. 2015).

State-of-the-art contributions to this domain include algorithms such as the Fitness Dependent Optimizer (FDO) (Abdullah and Ahmed 2019). It is inspired by the generative behaviours of bees, which adapt the agents' movements according to the fitness level of the agents to improve both convergence and solution quality. In the same way, the Ant Nesting Algorithm (ANA) (Rashid, Rashid, and Mirjalili 2021), It takes inspiration from ants' nesting behaviour intending to introduce new mechanisms for search agent updates that strike a better balance between exploration and exploitation. One more significant example is the FOX Optimization Algorithm (FOX) by (Mohammed and Rashid 2023). It emulates the foraging behaviour of foxes and balances global exploration and local exploitation to avoid local optima effectively.

The hybrid FOX-TSA algorithm combines the global search ability of the FOX algorithm with the fine-tuning strengths of the Tree-Seed Algorithm (TSA) (S.A. Aula and Rashid 2024). This combination creates a powerful tool that balances exploration and exploitation, making it ideal for improving optimization tasks like neural network training. Traditional methods such as Stochastic Gradient Descent (SGD) often struggle in complex optimization problems (Ruder 2016;Sharma 2018), but hybrid FOX-TSA helps overcome these limitations. Innovative nature-inspired algorithms (NIAs), including FDO, ANA, and FOX, play a key role in solving such challenges (Fadakar 2023; Nyandieka and Segera 2023). The hybrid FOX-TSA stands out by offering better performance and scalability, especially for machine learning applications (Sirwan A. Aula and Rashid 2024b).

## 1.2. Research Problem

Even with their common use, several optimizers, including the Adam and SGD, have severe limitations in suitability for complicated, multimodal, and constrained optimization problems (Talele and Phalnikar 2023). Adam is especially known for adjusting learning rates to non-stationary objectives, and for that reason is a strong candidate in the realm of optimization tasks. (Liu et al. 2020; Zhou, Huang, and Zhang 2022). Unfortunately, its performance degrades due to convergence to suboptimal solutions and sensitivity to hyperparameter setting (Ma, Wu, and Yu 2023; Li, Zhang, and Yoon 2021); (Li et al. 2020). However, unlike other methods, SGD remains as one of the most basic optimization techniques, and while it still is not adaptive it still needs plenty of careful hyperparameter tuning to reach optimal results (Aytaç, Güneş, and Ajlouni 2022) ; (Sathyabama and Saruladha 2022) ; (Mandt, Hof Fman, and Blei 2017); (Wu et al. 2021); (Wei et al. 2022).

However, these limitations become all the more apparent in large scale neural network and datasets with intricate patterns and dependencies, in which optimization stability and speed is essential first, and precision second. To address these challenges, we need to develop an advanced optimization framework to further promote neural network training in terms of robustness and efficiency. (Mehmood, Ahmad, and Whangbo 2023; Tang et al. 2018; Shrestha and Mahmood 2019; Cui et al. 2020;Aytaç, Güneş, and Ajlouni 2022).

This research presents Foxtsage, a hybrid optimization algorithm that combines the global exploration power of the FOX-TSA with the stability and adaptability of SGD to overcome these challenges. This approach aims to supplement the shortcomings of Adam and SGD with a better balance in the optimization framework which retains simplicity and computational stability. This work rigorously benchmarks Foxtsage across various datasets and metrics. Also investigates whether Foxtsage can reshape optimization for neural networks or not.

### 1.3. Objectives

This study has the following main objectives:

1. To apply the hybrid Foxtsage as an adaptive learning rate design in order to improve neural network training.
2. Benchmark datasets were used to compare the performance of the Foxtsage vs Adam optimizer.
3. To evaluate the essential performance criteria, i.e., the training loss, accuracy, and computing time, and experiment with statistical significance tests to verify the robustness of the Foxtsage .

### 1.4. Motivations

1. Although popular optimizers such as Adam and SGD face particular optimization challenges. However, Adam is sensitive to the hyperparameters and often converges to suboptimal solutions, hence its applicability is limited to certain optimization problems (Talele and Phalnikar 2023). SGD, like ADAM, basically needs a broad range of tuning to respond in the optimal direction (Sathyabama and Saruladha 2022; Mandt, Hofmann, and Blei 2017).
2. High capabilities to balance global exploration and local exploitation as well as strong potential to improve SGD make FOX-TSA an appropriate solution. Its hybrid nature is able to address the limitations of standalone algorithms in complex and multimodal tasks (Sirwan A. Aula and Rashid 2024)
3. Consequently, integrating FOX-TSA with SGD offers the possibility to evaluate if a hybrid approach can outperform Adam, one of the most widely adopted optimizers for neural network training (Kingma and Ba 2014). This research attempts to enhance the adaptability and convergence speed of SGD by taking advantage of FOX-TSA's strengths.
4. Real world applications, such as healthcare, finance and autonomous systems, require neural networks. Using advanced frameworks like Foxtsage directly affects the reliability and performance of such systems, thus, Foxtsage is of paramount importance to modern AI solutions (Shrestha and Mahmood 2019; Cui et al. 2020;Aytaç, Güneş, and Ajlouni 2022).
5. The advantage of this is that since SGD is often said to be Adam's closest competitor ((Liu et al. 2020; Zhou, Huang, and Zhang 2022), it provides a straightforward opportunity for direct performance comparison with an enhanced version of SGD. This task not only fills in the gaps in the existing techniques but also attempts to test Foxtsage's ability to either bridge the gap or set an entirely new benchmark in neural network optimization.

### 1.5. Contributions

This study made known to novel advancements to the field of neural network optimization, and concluded the following main contributions:

1. We leveraged the integration of FOX-TSA with SGD, and called it Foxtsage, to introduce a new optimization framework. We use FOX-TSA's adaptive capability to improve the performance of SGD, overcoming the typical bounds of non-adaptive optimizers by leveraging its adaptivity for global and local search.
2. We validate the effectiveness of the Foxtsage through comprehensive experimental results: with a 42.03% mean loss reduction and a 42.19% standard deviation reduction over Adam. In addition, Foxtsage shows small yet significant improvements (0.78% in accuracy mean, 0.91% in precision mean, 1.02% in recall mean and 0.89% in F1_score mean). These results especially highlight its enhanced predictive robustness and efficiency.
3. Adam optimizer is benchmarked against Foxtsage with several standard datasets. It is shown that Foxtsage surpasses all previous models by a wide margin in key metrics like training loss as well as robustness across the datasets with the cost of increased computation time as a trade-off.
4. This work identifies critical challenges, such as the computational cost of Foxtsage (a 330.87% increase in time mean) and the need for dataset-specific adaptations. These insights provide a foundation for future research aimed at improving time complexity and scalability while retaining optimization benefits.

By addressing existing gaps and demonstrating the potential of hybrid algorithms like Foxtsage, this research contributes a significant advancement to the literature on neural network optimization techniques.

### 1.6. Paper Structure

The remainder of this paper is organized as follows. Section 2 reviews related work on NIAs and neural network optimization. Section 3 details the methodology, including the integration of FOX-TSA with SGD. Section 4 presents experimental results and analysis, highlighting performance improvements. Section 5 discusses the findings, including limitations and practical implications. Finally, Section 6 concludes the study and outlines directions for future work.

## 2. Literature Review
### 2.1. Overview of Optimization Algorithms

Optimization algorithms are essential for improving machine learning models, by making training faster, increasing accuracy, and enhancing overall performance robustness (Yang and Shami 2020; Hamdia, Zhuang, and Rabczuk 2021). Among the most popular optimizers are SGD and Adam. SGD is simple and reliable but often needs careful tuning of parameters to perform well (Mandt, Hoffman, and Blei 2016; Smith et al. 2018). In contrast, Adam combines adaptive learning rates with momentum, making it a more versatile and robust choice for many tasks (Guo et al. 2022); (El-Shazli, Youssef, and Soliman 2022). However, these methods have limitations, such as susceptibility to local minima (SGD) or suboptimal generalization (Adam) (B. Zhang et al. 2021). Nature-inspired algorithms (NIAs), such as the Grey GWO, PSO, and TSA (Kiran 2015), have emerged as alternatives to traditional optimizers. These algorithms mimic biological, ecological, or physical systems to explore and exploit the solution space effectively. For example, GWO simulates the hunting strategy of grey wolves, PSO models social behaviours like bird flocking, and TSA emulates the growth and dispersal patterns of trees (Gharehchopogh 2022).

## 2.2. Previous Studies

The literature has extensively explored the application of NIAs in optimization tasks. For example:

- The hybrid FOX-TSA algorithm leverages the exploration capabilities of the FOX algorithm and the exploitation strengths of TSA. Studies show it outperforms standalone FOX and TSA as well as traditional methods like GWO and PSO on CEC benchmark functions and engineering design problems (Sirwan A. Aula and Rashid 2024; Kiran 2015; Mohammed and Rashid 2023) .
- The Adam optimizer has been applied across diverse domains, such as natural language processing, computer vision, and predictive modeling. For instance, Adam exhibited competitive performance on the MNIST dataset but struggled with generalization in some tasks (Kingma and Ba 2014).
- Hybrid optimization methods, combining NIAs with conventional techniques, have demonstrated significant improvements. For example, the FOX-TSA hybrid has been applied to tourism datasets, achieving superior accuracy (94.64%) and faster convergence, reducing iterations by 25% (Sirwan A. Aula and Rashid 2024).

## 2.3. Research Gaps

Despite the demonstrated efficacy of hybrid algorithms like FOX-TSA, their integration with widely used optimizers, such as SGD or Adam, remains underexplored. Specifically, no direct comparisons between a Foxtsage optimizer and Adam for neural network training exist. Additionally, the impact of FOX-TSA on computational efficiency and time complexity in real-world applications warrants further investigation (Sirwan A. Aula and Rashid 2024).

## 2.4. Applications and Performance Comparisons

The following table summarizes the performance of Adam compared to other algorithms in various studies:

| Algorithm | Study/Application | Key Findings/Comparison with Adam |
|---|---|---|
| Adam(Kingma and Ba 2014) | Neural network training | Reliable convergence; limitations in generalization on complex datasets. |
| FOX-TSA (Sirwan A. Aula and Rashid 2024) | Predictive modeling, optimization | Superior accuracy (94.64%), and faster convergence (25% fewer iterations). |
| PSO(Kennedy, Eberhart, and gov 1995) | Demand forecasting, route planning | Effective in improving prediction accuracy but prone to local optima. |
| GWO (Mirjalili, Mirjalili, and Lewis 2014) | Resource optimization, planning | Robust against local optima; struggles with high-dimensional search spaces. |

| Algorithm | Study/Application | Key Findings/Comparison with Adam |
|---|---|---|
| Hybrid ABC (Forouzandeh, Rostami, and Berahmand 2022) | Recommender systems | Enhanced recommendation quality; lacked exploration of other algorithms. |
| Chaotic Whale Optimization (Kaur and Arora 2018) | Resource planning, data modeling | Improved exploration and convergence through chaotic dynamics. |
| Stochastic Gradient Descent | (Soydaner 2020) | Adam demonstrated faster convergence and achieved higher accuracy than SGD across multiple datasets. |
| RMSProp | (De, Mukherjee, and Ullah 2018) | Adam showed superior performance in terms of convergence speed and final accuracy compared to RMSProp. |
| AdaGrad | (Defossez et al. 2020; Desai 2020) | Adam outperformed AdaGrad by maintaining a more consistent learning rate, leading to better generalization. |
| HN_Adam | (Reyad, Sarhan, and Arafa 2023) | HN_Adam, a modified version of Adam, achieved higher accuracy and faster convergence than standard Adam on MNIST and CIFAR-10 datasets. |
| Adafactor | (Luo et al. 2023) | Adafactor performed comparably to Adam in language modeling tasks, with similar optimal performance and hyperparameter stability. |
| Lion | (Zhao et al. 2024) | Lion's performance was on par with Adam in terms of convergence and final model accuracy in language modeling. |
| Extended Kalman Filter (EKF) | (Shao et al. 2021; Vural, Ergut, and Kozat 2021); (Tripathi et al. 2023) | EKF-trained neural networks exhibited better transferability and were less sensitive to learning rate variations compared to those trained with Adam. |
| Levenberg-Marquardt (LM) | (Khatti and Grover 2024; Manohar and Das 2022) | The LM algorithm achieved rapid convergence to machine precision, outperforming Adam in training efficiency for small to medium-sized neural network |

The literature highlights Adam's robustness in certain scenarios but also underscores its limitations in handling complex, multimodal optimization problems. Comparatively, FOX-TSA exhibits superior performance metrics, such as improved predictive accuracy and reduced iterations, particularly in hybrid applications (Sirwan A. Aula and Rashid 2024; Sirwan A. Aula and Rashid 2024).

## 3. Methodology

This section outlines the experimental methodology adopted to evaluate the effectiveness of Foxtsage with SGD compared to the Adam optimiser. The methodology comprises four subsections: the neural network architectures used in experiments, an overview of the optimisers, the benchmark datasets employed, and the experimental setup.

## 3.1. Neural Network Architecture

For our experiments, we utilized two primary neural network architectures: ConvNet and Multi Layer Perceptrons. The suitability of the selected architectures for handling the characteristics of the benchmark datasets is the reason that they were chosen:

1. Convolutional Neural Networks are well suitable for image-based datasets like MNIST and CIFAR-10 because of the auto learning of spatial hierarchies and feature extraction like edges, textures and shapes (Watanabe and Yamana 2022; Hirata and Takahashi 2023; H. Zhang and Ma 2020; Al Badawi et al. 2021). A CNN with multiple convolutional and pooling layers and a fully connected layer was trained for the CIFAR-10 dataset. Similar to the CNN for CIFAR, the CNN for MNIST but with fewer parameters as a simpler dataset (MNIST) to work on.
2. For example, Multi-Layer Perceptrons were used in benchmark problems like logistic regression, tabular datasets or even video inpainting (Khashei and Hajirahimi 2019 ; Assi et al. 2018; Islam et al. 2022; Watanabe and Yamana 2022; Hirata and Takahashi 2023; H. Zhang and Ma 2020). Fully connected networks show themselves to be versatile and effective for structured data. The MLP architecture consisted of an input layer, two hidden layers with ReLU activation function between layers and an output layer.

The architecture choice made optimisers compatible with the dataset's characteristics and a fair comparison across different tasks.

## 3.2. Optimizers Overview
### 3.2.1 Adam Optimizer

Adam is an adaptive optimization algorithm that adjusts the learning rate of each parameter dynamically during training. It combines the benefits of AdaGrad and RMSProp by leveraging estimates of the first and second moments of gradients. This makes Adam robust to noisy gradients and well-suited for non-convex optimization problems. The learning rate in Adam is updated as follows (Kingma and Ba 2014):

$$m_t = \beta_1 m_{t-1} + (1 - \beta_1) g_t \tag{1}$$

$$v_t = \beta_2 v_{t-1} + (1 - \beta_2) g_t^2 \tag{2}$$

$$\widehat{m_t} = \frac{m_t}{1 - \beta_1^t}, \quad \widehat{v_t} = \frac{v_t}{1 - \beta_2^t} \tag{3}$$

$$\theta_{t+1} = \theta_t - \eta \frac{\widehat{m_t}}{\sqrt{\widehat{v_t}} + \epsilon} \tag{4}$$

Where $\eta$ is the learning rate, $\beta 1$ and $\beta 2$ are the exponential decay rates for the first and second moment estimates respectively, $\boldsymbol{g_t}$ is the gradient of the objective function with respect to the parameters at time $t, and$ $\epsilon$ is a small constant added for numerical stability to prevent division by zero. $\boldsymbol{m_t}$ $and$ $\boldsymbol{v_t}$ are the biased first and second-moment estimates, corrected by $\widehat{m_t}$ and $\widehat{v_t}$, which account for bias introduced in early timesteps due to initialization. $\theta_t$ represents the model parameters at time $t$, updated iteratively to $\theta_{t+1}$ using the adjusted gradient values.

### 3.2.2 Stochastic Gradient Descent

SGD updates model parameters iteratively by computing the gradient of the loss function for a randomly selected mini-batch. Despite its simplicity and computational efficiency, SGD often requires careful tuning of the learning rate ($\eta$) to achieve convergence (Mandt, Hof Fman, and Blei 2017):

$$\theta_{t+1} = \theta_t - \eta \nabla_{\theta \mathcal{L}}(\theta_t) \tag{5}$$

Here, $\nabla_{\theta \mathcal{L}}$ represents the gradient of the loss function. However, the static nature of $\eta$ makes SGD prone to convergence issues in complex optimization landscapes.

### 3.2.3 Foxtsage: Hybrid FOX-TSA with SGD

The Foxtsage optimiser, a hybrid integration of the FOX-TSA algorithm with SGD, was employed to dynamically update the learning rate $\eta$, addressing the limitations of static learning rates. Foxtsage combines the exploration capabilities of the FOX optimizer to effectively navigate the search space and the exploitation strengths of the Tree-Seed Algorithm for fine-tuned convergence. At each iteration, Foxtsage adaptively adjusts the learning rate based on the best solution identified by the Hybrid FOX-TSA. This dynamic approach enables SGD to achieve both stability and accelerated convergence, reducing the risk of premature stagnation and enhancing neural network training performance.

$$\eta_t = \eta_{\text{base}} \cdot \frac{1}{1 + \alpha \cdot f_{\text{best}}(t)} \tag{6}$$

Where $\eta_{\text{base}}$ is the initial learning rate, $\alpha$ is a decay control parameter and $f_{\text{best}}(t)$ is the best fitness value at iteration $t$.

This dynamic adjustment ensures better convergence, balances exploration and exploitation, and avoids the pitfalls of static learning rates.

Pseudocode for Foxtsage Optimizer:

**Algorithm: Foxtsage - Hybrid FOX-TSA with SGD Optimizer**

1. Initialize:
   a. Randomly initialize a population of learning rates:
      population ← Uniform($\eta\_min, \eta\_max, P$)
   b. Set best_lr ← population[0]   # Initial best learning rate
   c. Set best_loss ← ∞          # Initialize best loss to a high value

2. FOR iteration = 1 to I:   # Iterate for the given number of iterations
   a. FOR each candidate_lr ∈ population:
      i. Initialize optimizer SGD($\theta$, candidate_lr)
      ii. Train the model $\theta$ on D_train for one epoch:
         $\theta \leftarrow \theta$ - candidate_lr × $\nabla L(\theta)$
      iii. Compute loss:
         current_loss ← $L(\theta)$
      iv. IF current_loss < best_loss:
            Update best solution:
               best_loss ← current_loss
               best_lr ← candidate_lr

   b. Update population using FOX-TSA mechanism:
      FOR each candidate_lr ∈ population:
         i. Generate a random value r ∈ [0, 1]
         ii. IF r < 0.5:  # Exploration step
               candidate_lr ← best_lr × (1 + Random_Gaussian())
            ELSE:       # Exploitation step
               candidate_lr ← best_lr / (1 + Random_Gaussian())
         iii. Clip candidate_lr to the bounds [$\eta\_min, \eta\_max$]:
            candidate_lr ← Clip(candidate_lr, $\eta\_min, \eta\_max$)

   c. Print performance metrics for the current iteration:
      Print: "Iteration:", iteration, "Best Loss:", best_loss,
         "Best Learning Rate:", best_lr

3. Return:
   a. $\theta^*$: Final optimized model parameters
   b. $\eta^*$: Best learning rate (best_lr)

Summary of Foxtsage Pseudocode: The Foxtsage optimiser integrates the global exploration of the Hybrid FOX-TSA with the local convergence of SGD. It dynamically adjusts the learning rate to enhance model training efficiency.

1. Initialization:
    o A population of candidate learning rates is randomly initialized within specified bounds ($\eta\_min, \eta\_max$)
2. Training with SGD:
    o For each learning rate in the population, the model is trained for one epoch using SGD, and the corresponding loss is evaluated.
    o The best learning rate (best_lr) is updated based on the lowest loss.

3. Learning Rate Update (FOX-TSA Mechanism):
    - Learning rates are updated dynamically using FOX-TSA principles:
        - Exploration: Introduce randomness to increase diversity.
        - Exploitation: Fine-tune around the best learning rate.
    - Updated values are clipped to stay within the predefined range.
4. Iteration:
    - The process repeats for a fixed number of iterations, improving the learning rate at each step.
5. Output:
    - The final optimized model parameters and the best learning rate are returned.

### 3.3. Benchmark Datasets

We used the following benchmark datasets:

1. MNIST: A dataset of handwritten digits, consisting of 60,000 training images and 10,000 testing images, each of size $28 \times 28$.
2. CIFAR-10: A dataset of 60,000 $32 \times 32$ colour images categorized into 10 classes.
3. IMDB: A textual dataset used for binary sentiment classification.
4. Logistic Regression Variants: Structured datasets to evaluate the performance of logistic regression models.

For image datasets, normalization was applied to scale pixel values to the range [0, 1], and one-hot encoding was used for the target labels. For the IMDB dataset, text pre-processing included tokenization, removal of stop words, and padding for uniform sequence lengths. Then the data was split into 80% for training and 20% for testing for all datasets.

### 3.4. Experimental Setup

#### 3.4.1 Computing Environment

The experiments were conducted on a high-performance GPU environment using Kaggle and Google Colab. The hardware included NVIDIA Tesla GPUs with CUDA support. The software stack comprised TensorFlow, PyTorch, and custom implementations of Foxtsage in Python.

#### 3.4.2 Settings and Hyperparameters

Two experimental settings were defined to evaluate the performance of the optimizers:

1. Setting 1: 5 iterations with a population size of 10 agents.
2. Setting 2: 50 iterations with a population size of 30 agents.

Hyperparameters for FOX-TSA included exploration and exploitation weights, while SGD used a base learning rate of $\eta_{base} = 0.01$. Adam followed the standard settings: $\beta_1 = 0.9, \beta_2 = 0.999, \epsilon = 10^{-8}$.

#### 3.4.3 Evaluation Metrics

The performance was evaluated using the following metrics:

- Loss Mean: Average loss across the test set.

$$\mathcal{L}(\theta) = -\frac{1}{N}\sum_{i=1}^{N}[y_i \log(\hat{y}_i) + (1-y_i)\log(1-\hat{y}_i)] \qquad (7)$$

Where: $\mathcal{L}(\theta)$ represents the binary cross-entropy loss computed over the model parameters $\theta$, $N$ is the total number of samples in the test set, $i$ is the index of a specific sample ranging from 1 to $N$, $y_i$ is the true label for the $i$-th sample (either 0 or 1 in binary classification), $\hat{y}_i$ is the predicted probability (output of the model) for the $i$-th sample belonging to the positive class (1), and $\log(\hat{y}_i)$ denotes the natural logarithm function. This loss function measures the average negative log-likelihood of the predictions, penalizing incorrect predictions to improve the model's accuracy

- Accuracy Mean: Proportion of correctly classified instances.

$$\text{Accuracy} = \frac{\text{Number of Correct Predictions}}{\text{Total Number of Predictions}} \qquad (8)$$

- F1-Score: Harmonic mean of precision and recall.

$$F1 = 2 \cdot \frac{\text{Precision} \cdot \text{Recall}}{\text{Precision} + \text{Recall}} \qquad (9)$$

- Computation Time: Average time required for training in seconds.

$$T_{avg} = \frac{1}{N}\sum_{i=1}^{N} T_i \qquad (10)$$

Where $T_{avg}$ is the average computation time in seconds, $N$ is the total number of runs or iterations, and $T_i$ is the computation time for the $i$-th run or iteration.

## 4. Results and Discussion

In this section, we provide a detailed analysis of the results obtained using the Hybrid FOX-TSA with SGD and Adam optimisers across various datasets and models. The results are presented in terms of training loss and accuracy, comparing both optimisers under two experimental settings: 5 iterations, 10 population size (Setting 1) and 50 iterations, 30 population size (Setting 2).

## 4.1. Results
### 4.1.1 Logistic Regression

In the framework of MNIST logistic regression, we showed a comparative analysis using the Foxtsage algorithm and the Adam optimizer across two distinct experimental setups: Setting 1 and Setting 2. These settings were the only ones that differed mostly in their iteration count and population size, which gave a more extensive exploration of the optimization landscape, compared to Setting 2, which had a higher iteration count and a larger population.

In Setting 1, we show that the Foxtsage algorithm outperforms the Adam optimizer on a set of all key metrics. Foxtsage was also able to achieve a much lower mean loss (0. 2891) compared to Adam (0.3010), which demonstrates that Foxtsage is good at reducing loss and optimising model parameters. Furthermore, Foxtsage achieved a mean accuracy of 91.81%, which was higher than Adam's accuracy of 91.33%. The F1-score also reflected this trend, with Foxtsage attaining a mean value of 0.9179, while Adam achieved 0.9131. Additionally, the variability in performance, as indicated by the standard deviation of the metrics, was lower for Foxtsage. For instance, the loss standard deviation for Foxtsage was 0.0153, compared to 0.0179 for Adam, suggesting more consistent performance during optimization. These observations are supported by (**Table 1:** MNIST logistic regression: Tabular comparison of loss and accuracy for Setting 1 and Figures **Figure 1:** MNIST logistic regression: Loss comparison graph for Setting 1 and Figure 2: MNIST logistic regression: Accuracy comparison graph for Setting 1), which depicts the tabular and graphical comparisons of loss and accuracy in Setting 1.

*Table 1*: *MNIST logistic regression: Tabular comparison of loss and accuracy for Setting 1*

| Model | Loss Mean | Loss StdDev | Accuracy Mean | Accuracy StdDev | F1-Score Mean | F1-Score StdDev |
|---|---|---|---|---|---|---|
| Foxtsage | 0.289095 | 0.015273 | 0.91812 | 0.002988 | 0.917886 | 0.003024 |
| Adam | 0.301014 | 0.017938 | 0.91334 | 0.005242 | 0.913074 | 0.005383 |

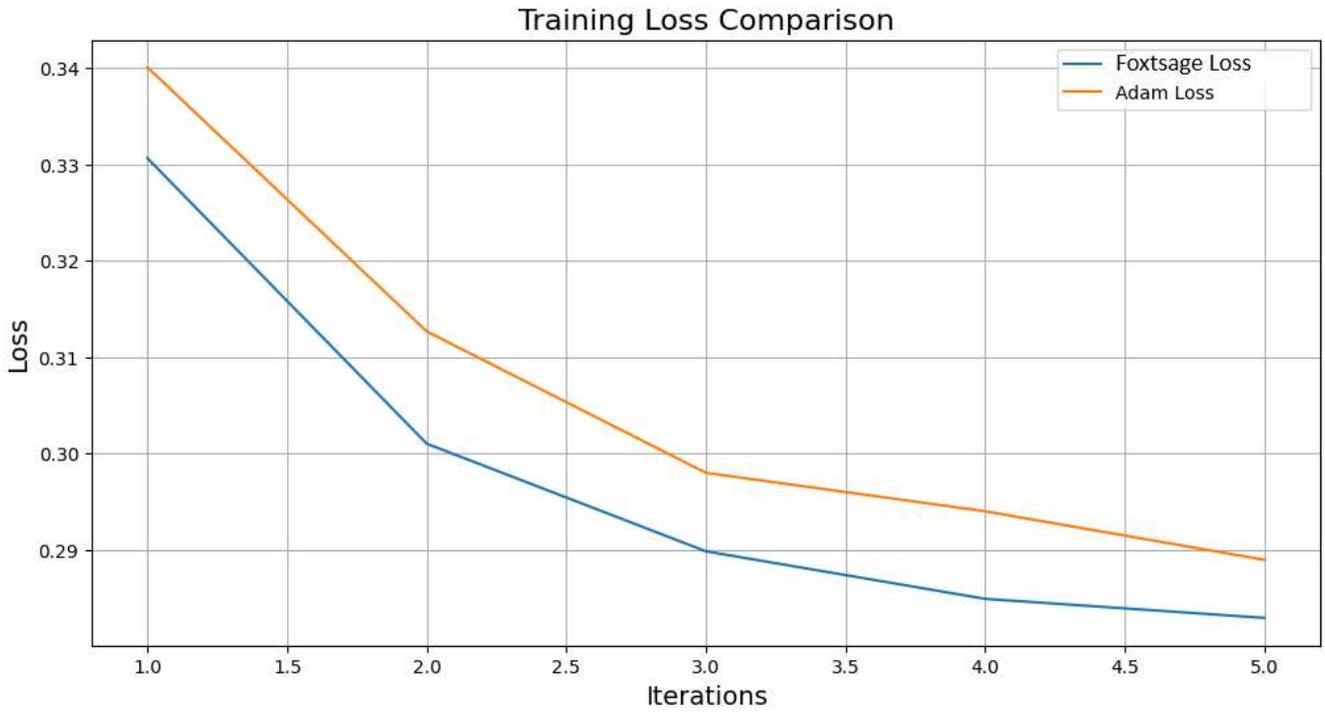

*Figure 1: MNIST logistic regression: Loss comparison graph for Setting 1*

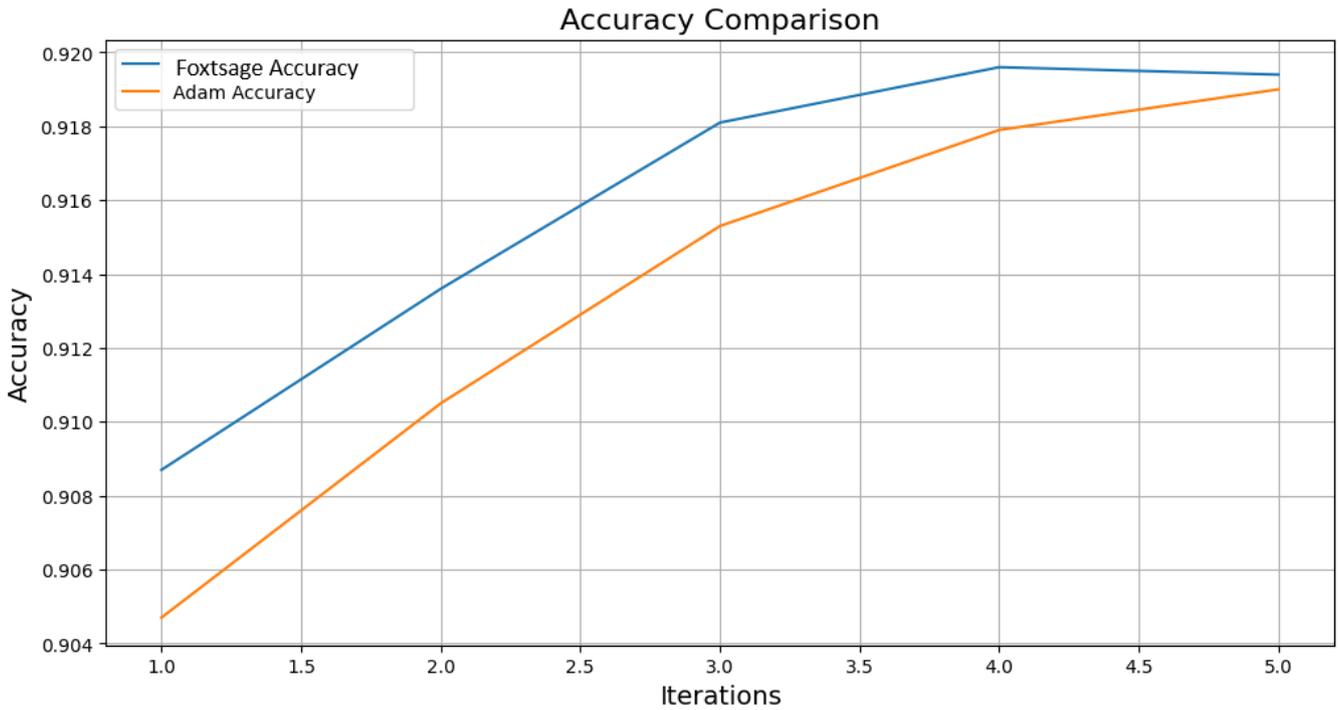

*Figure 2: MNIST logistic regression: Accuracy comparison graph for Setting 1*

Setting 2, characterized by enhanced computational resources in terms of iterations and population size, further highlighted the superiority of Foxtsage . Under these conditions, Foxtsage reduced the mean loss to 0.2686, while Adam achieved a loss of 0.2819. This reduction underscored the Foxtsage's ability to explore and exploit the search space more effectively with increased resources. Accuracy also improved in Setting 2, with Foxtsage achieving a mean accuracy of 92.27%, surpassing Adam's performance of 91.99%. The F1-score followed a similar pattern, with Foxtsage achieving a mean value of 0.9225, compared to Adam's 0.9198. Notably, the standard deviation values for loss, accuracy, and F1-score were markedly lower for Foxtsage in this setting. For instance, the loss standard deviation was reduced to 0.0021 for Foxtsage, compared to 0.0117 for Adam, indicating greater stability and reliability in performance. These findings are illustrated in (**Table 2**: MNIST logistic regression: Tabular comparison of loss and accuracy for Setting 2) and **Figures (Figure 3**: MNIST logistic regression: Loss comparison graph for Setting 2 and **Figure 4**: MNIST logistic regression: Accuracy comparison graph for Setting 2), which provide detailed comparisons of loss and accuracy metrics for Setting 2.

***Table 2**: MNIST logistic regression: Tabular comparison of loss and accuracy for Setting 2*

| Model | Loss Mean | Loss StdDev | Accuracy Mean | Accuracy StdDev | F1-Score Mean | F1-Score StdDev |
|---|---|---|---|---|---|---|
| Foxtsage | 0.26856156 | 0.002089195 | 0.92274 | 0.000613122 | 0.922542424 | 0.000616471 |
| Adam | 0.281870654 | 0.011731651 | 0.919964 | 0.004239935 | 0.919797728 | 0.004220538 |

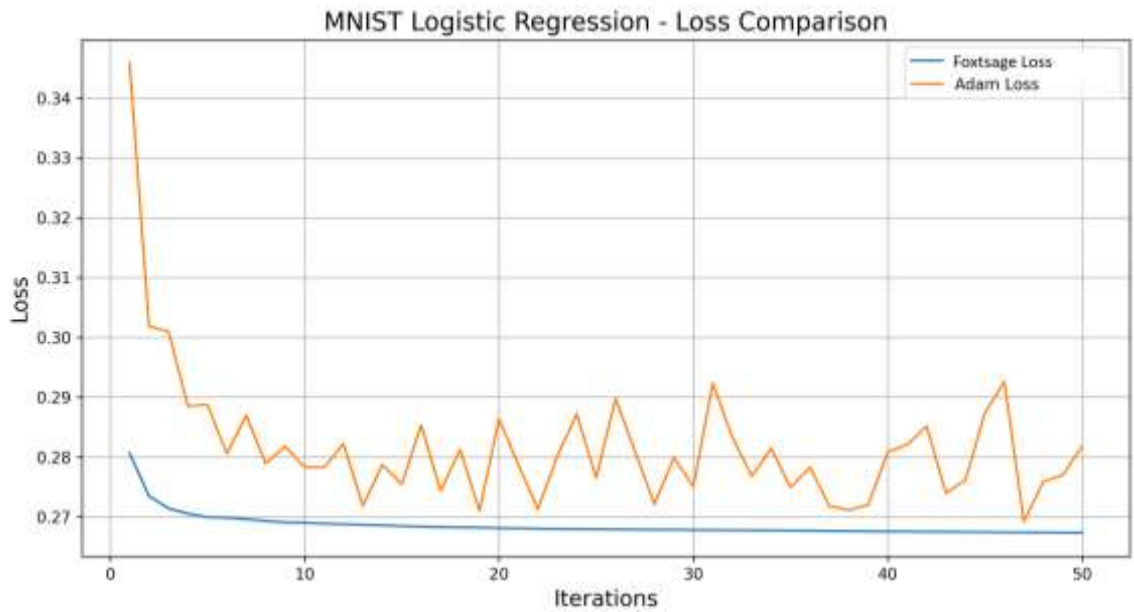

*Figure 3: MNIST logistic regression: Loss comparison graph for Setting 2*

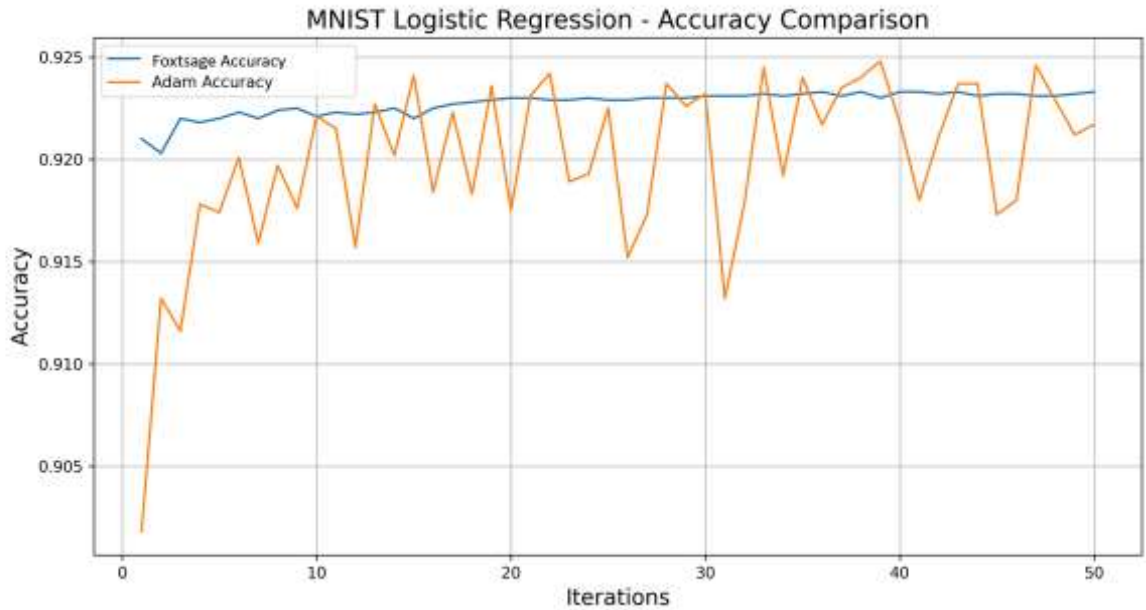

*Figure 4: MNIST logistic regression: Accuracy comparison graph for Setting 2*

Across both experimental setups, Foxtsage consistently outperformed Adam in terms of lower training loss, higher accuracy, and superior F1-score. While Adam displayed faster initial convergence in Setting 1, the Foxtsage demonstrated better optimization over time, particularly in the more resource-intensive Setting 2. The improvements observed in Setting 2 highlight the benefits of increasing iterations and population size for Foxtsage, allowing it to capitalize on its robust optimization capabilities. The results of these findings are visually visible from the graphs and tables provided, as they allow for a detailed study of the comparative performance of the two algorithms under different conditions. These results confirm that Foxtsage can achieve consistent gains in MNIST logistic regression, especially in accuracy and loss reduction. Additionally, the algorithm's stability is reflected in lower standard deviations in favour of wider applications in neural network training.

### 4.1.2 Logistic Regression

For the baseline performance on the IMDB dataset, Foxtsage gave a significant benefit relative to Adam in both settings. In setting 1 fewer iterations and smaller population sizes were used. The Foxtsage achieved a lower mean loss of 0.3446 with a standard deviation of 0.0366, outperforming Adam, which recorded a mean loss of 0.3840 and a standard deviation of 0.0214 (**Table 3**: IMDB Logistic Regression: Tabular comparison of loss and accuracy for Setting 1). This indicates better convergence properties for the Foxtsage under constrained computational settings. In terms of accuracy, the Foxtsage reached a mean accuracy of 87.15% with a standard deviation of 1.28%, while Adam achieved 85.82% with a standard deviation of 0.24%. The F1-score further highlighted the superiority of the Foxtsage, with a mean score of 0.8757 compared to Adam's 0.8625 (**Figure 5**: Training Loss Comparison for IMDB Logistic Regression (Setting 1) and **Figure 6**: Accuracy Comparison for IMDB Logistic Regression (Setting 1)).

*Table 3: IMDB Logistic Regression: Tabular comparison of loss and accuracy for Setting 1*

| Model | Loss Mean | Loss StdDev | Accuracy Mean | Accuracy StdDev | F1-Score Mean | F1-Score StdDev |
|---|---|---|---|---|---|---|
| Foxtsage | 0.344642 | 0.036642 | 0.87152 | 0.012797 | 0.875703 | 0.012634 |
| Adam | 0.383995 | 0.021405 | 0.85828 | 0.002435 | 0.862502 | 0.001853 |

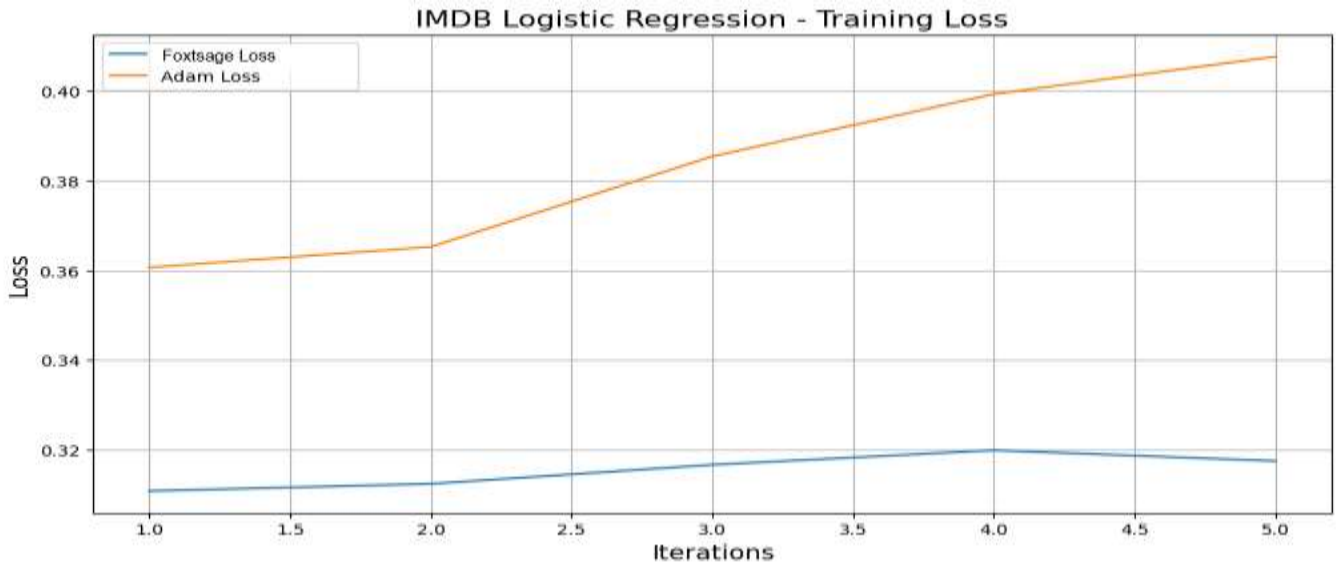

*Figure 5: Training Loss Comparison for IMDB Logistic Regression (Setting 1)*

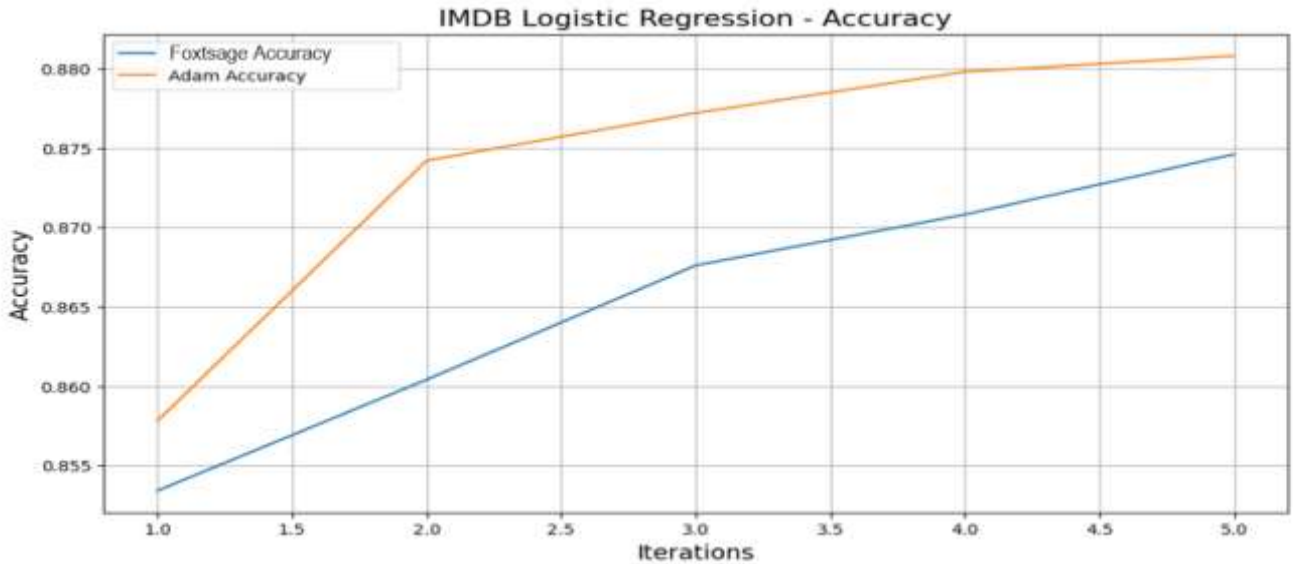

*Figure 6: Accuracy Comparison for IMDB Logistic Regression (Setting 1)*

Setting 2 leveraged higher iterations and a larger population size, allowing for more exploration of the optimization space. The Foxtsage continued to outperform Adam, reducing the mean loss to 0.2945 with a minimal standard deviation of 0.0067. In comparison, Adam's loss mean was 0.4879 with a standard deviation of 0.0387, showcasing the Foxtsage's robust convergence under more computationally intensive settings. For accuracy, the Foxtsage achieved 88.06% with a standard deviation of 0.14%, whereas Adam remained at 85.82% with a standard deviation of 0.31%. The F1-score of the Foxtsage also increased to 0.8806, outperforming Adam's 0.8582 see (**Table 4**: IMDB Logistic Regression: Tabular comparison of loss and accuracy for Setting 2).

*Table 4: IMDB Logistic Regression: Tabular comparison of loss and accuracy for Setting 2*

| Model | Loss Mean | Loss StdDev | Accuracy Mean | Accuracy StdDev | F1-Score Mean | F1-Score StdDev |
|---|---|---|---|---|---|---|
| Foxtsage | 0.294519251 | 0.006660367 | 0.8806 | 0.001361372 | 0.880554488 | 0.001356127 |
| Adam | 0.487864644 | 0.038650882 | 0.858243478 | 0.003092044 | 0.858210229 | 0.003099369 |

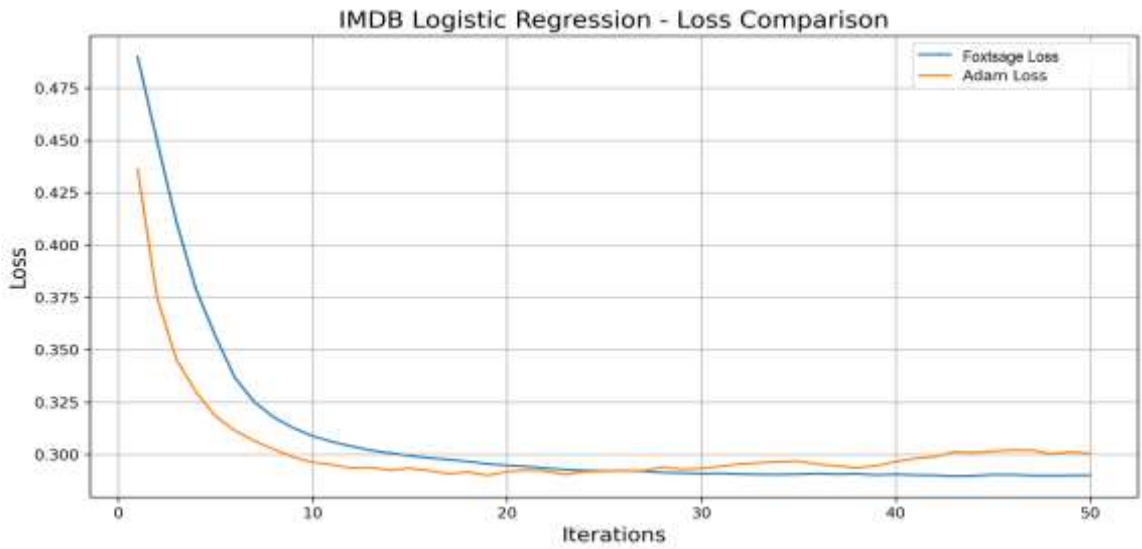

*Figure 7: Training Loss Comparison for IMDB Logistic Regression (Setting 2)*

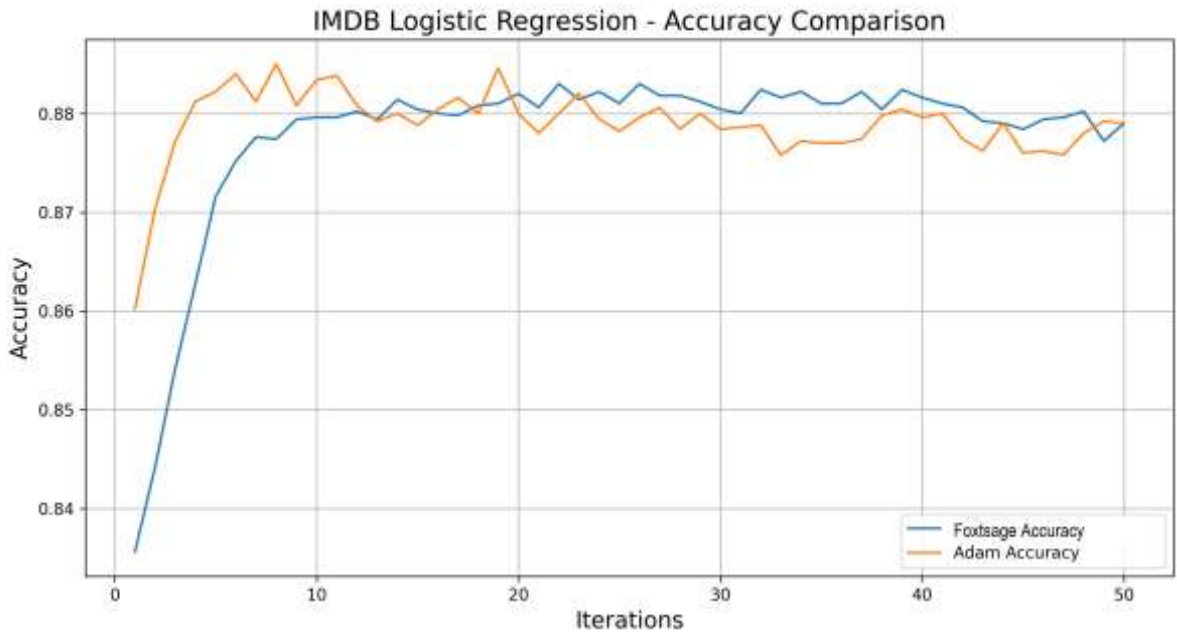

*Figure 8: Accuracy Comparison for IMDB Logistic Regression (Setting 2)*

The loss comparison graphs (**Figure 5, Figure 7**) for both settings, displayed above, clearly illustrate the more rapid and stable convergence of the Foxtsage optimiser compared to Adam. Similarly, the accuracy graphs (**Figure 6, Figure 8**) depict the Foxtsage consistently outperforming Adam in both settings, reflecting its ability to generalize better across iterations. These findings underscore the effectiveness of the Foxtsage optimizer in leveraging its dynamic adaptation of learning rates to outperform Adam in terms of both optimization efficiency and model performance on the IMDB dataset.

### 4.1.3 Multilayer Perceptron

The Multilayer Perceptron model trained on the MNIST dataset provides a significant perspective on the comparative performance of the Foxtsage optimizer and the Adam optimizer under both experimental settings. In Setting 1, characterized by fewer iterations and smaller population size, the Foxtsage demonstrated superior performance over the Adam optimizer in terms of loss and accuracy. Specifically, the Foxtsage achieved a mean training loss of 99.45 with a standard deviation of 35.84, while the Adam optimizer's training loss was significantly higher at 156.57 with a standard deviation of 66.97. This indicates a stronger optimization capability of the Foxtsage , especially under constrained settings. In terms of accuracy, the Foxtsage achieved a mean accuracy of 95.55% compared to Adam's 95.01%, showcasing a noticeable improvement. Additionally, the F1-score for the Foxtsage was 0.9554, marginally surpassing Adam's score of 0.9499. These results highlight the Foxtsage's ability to achieve better overall model performance in fewer iterations see (**Table 5**: MNIST MLP: Performance Metrics for Setting 1 (Loss, Accuracy, F1-Score).

*Table 5: MNIST MLP: Performance Metrics for Setting 1 (Loss, Accuracy, F1-Score)*

| Model | Loss Mean | Loss StdDev | Accuracy Mean | Accuracy StdDev | F1-Score Mean | F1-Score StdDev |
|---|---|---|---|---|---|---|
| Foxtsage | 99.4514 | 35.842947 | 0.9555 | 0.015802 | 0.95542 | 0.015931 |
| Adam | 156.57302 | 66.975898 | 0.95008 | 0.013938 | 0.94996 | 0.014041 |

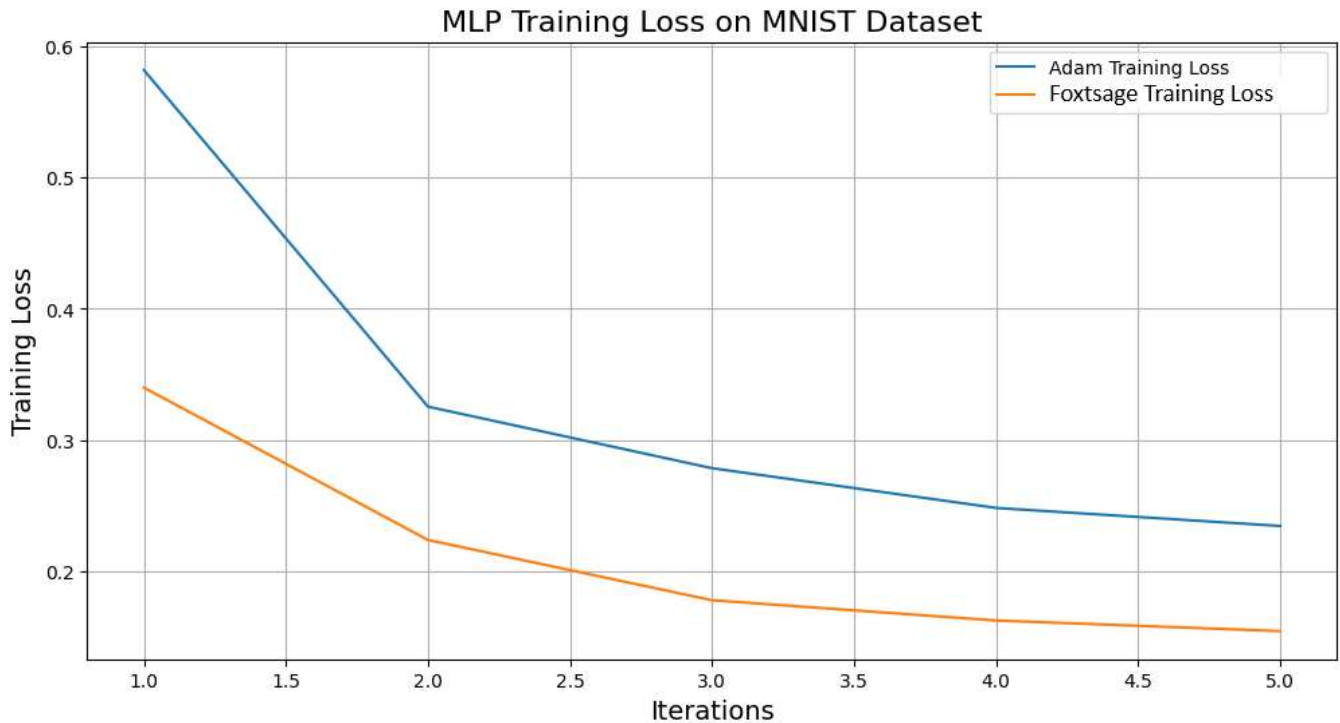

*Figure 9: MNIST MLP: Training Loss Comparison Graph (Setting 1)*

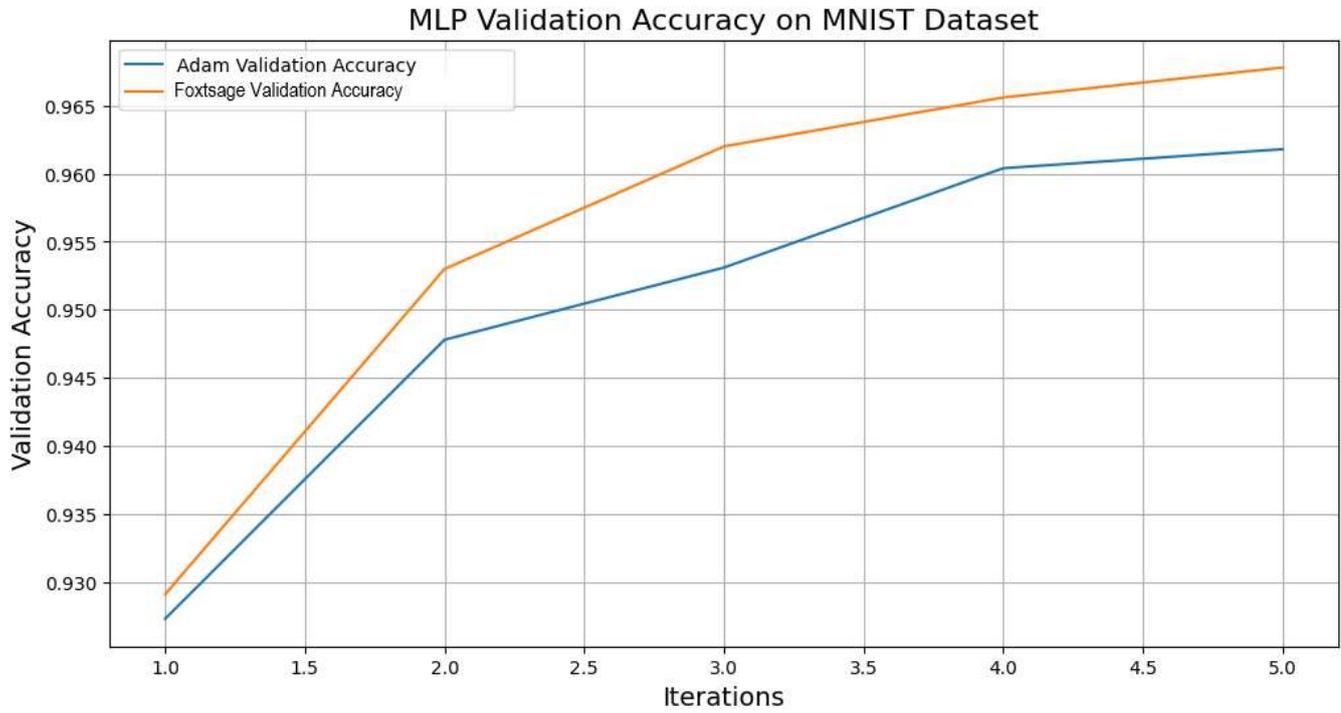

*Figure 10: MNIST MLP: Validation Accuracy Comparison Graph (Setting 1)*

Under Setting 2, with a larger number of iterations and a greater population size, the Foxtsage continued to outperform the Adam optimizer across all metrics. The mean training loss for the Foxtsage was reduced to 46.82 with a standard deviation of 19.55, while Adam's training loss, although reduced from Setting 1, remained higher at 80.95 with a standard deviation of 34.20. This demonstrates the Foxtsage's enhanced convergence and optimization efficiency with an increased computational budget. The accuracy achieved by the Foxtsage in Setting 2 was 97.69%, significantly higher than Adam's 97.14%. Similarly, the F1-score for the Foxtsage was 0.9768, which outperformed Adam's score of 0.9714 see (**Table 6**: MNIST MLP: Performance Metrics for Setting 2 (Loss, Accuracy, F1-Score)). The consistent improvement across metrics highlights the robustness of the Foxtsage optimizer in adapting to increased iterations and population sizes to deliver superior results.

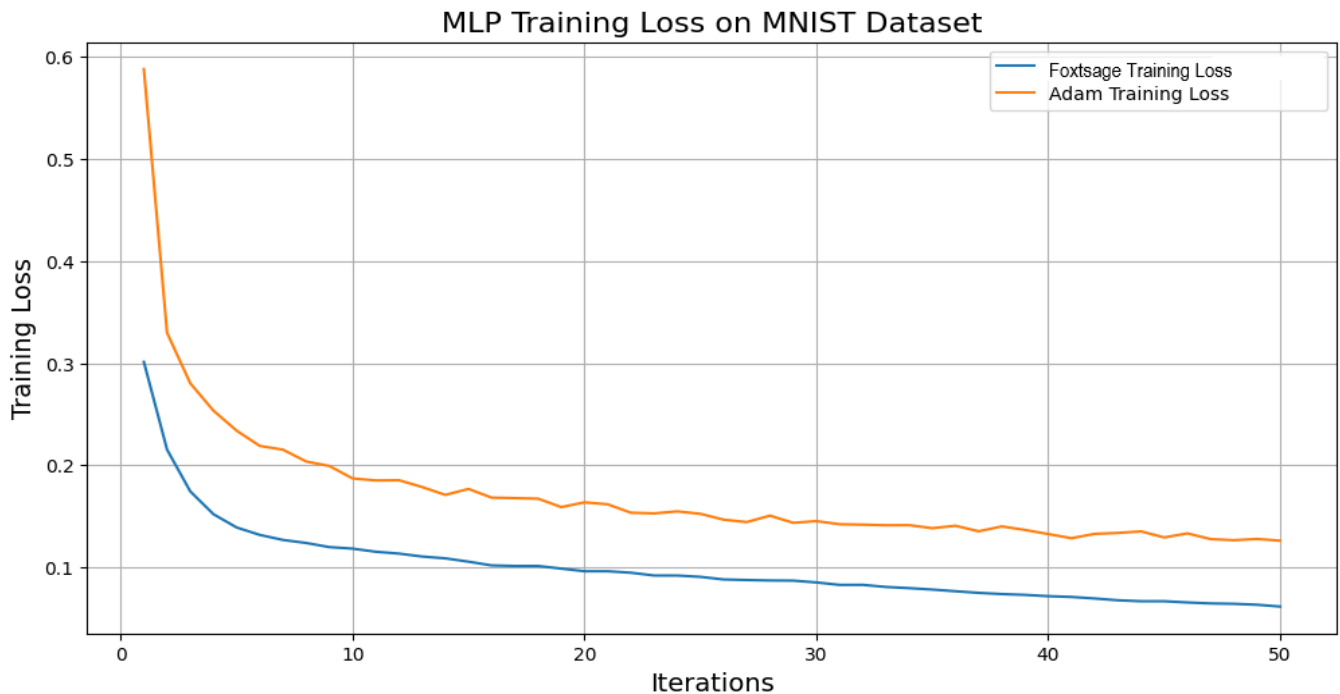

*Figure 11: MNIST MLP: Training Loss Comparison Graph (Setting 2)*

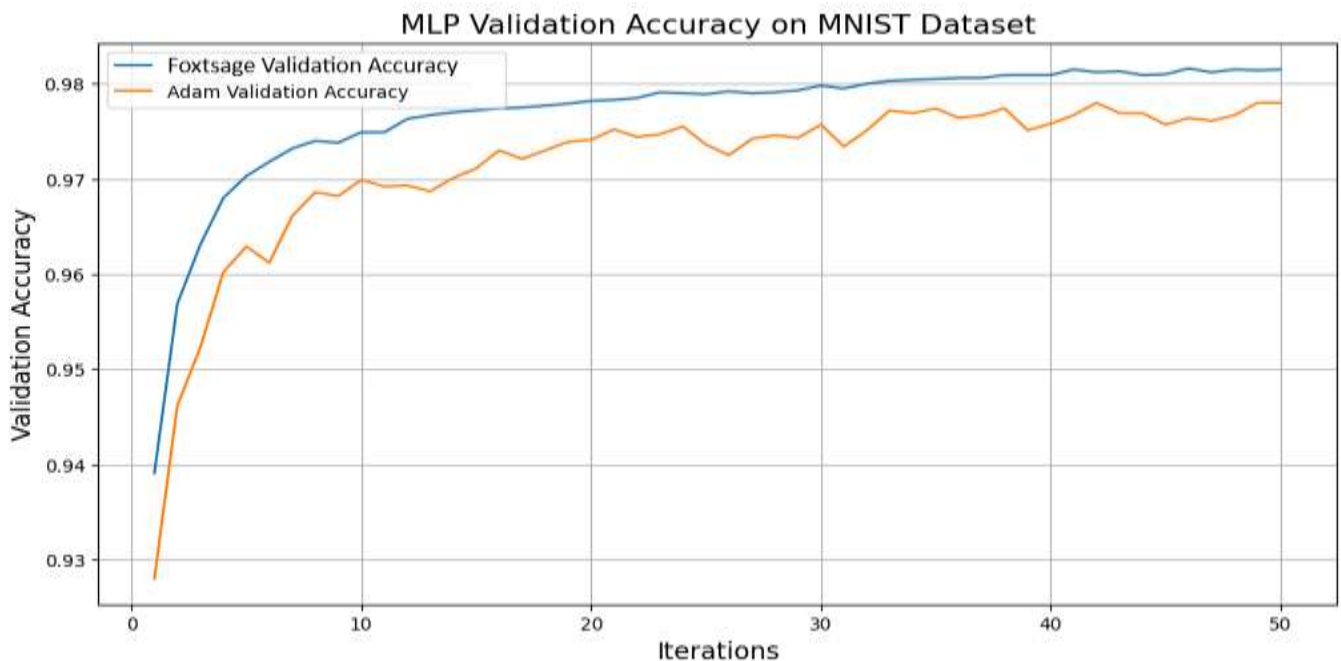

*Figure 12: MNIST MLP: Validation Accuracy Comparison Graph (Setting 2)*

The comparison between the two settings reveals that the Foxtsage not only achieves lower loss and higher accuracy but also maintains stability across iterations, as evidenced by its lower standard deviations in loss and accuracy see above loss graphs (**Figure 9, Figure 11**) and accuracy graphs (**Figure 10, Figure 12**). The Adam optimizer, while competitive, falls short of the Foxtsage's precision in parameter optimization, especially when the computational budget is increased.

***Table 6***: *MNIST MLP: Performance Metrics for Setting 2 (Loss, Accuracy, F1-Score)*

| Model | Loss Mean | Loss StdDev | Accuracy Mean | Accuracy StdDev | F1-Score Mean | F1-Score StdDev |
|---|---|---|---|---|---|---|
| Foxsage | 46.822948 | 19.55569392 | 0.976876 | 0.007259267 | 0.976872 | 0.007275512 |
| Adam | 80.951882 | 34.20677034 | 0.971472 | 0.008950659 | 0.971454 | 0.008995741 |

Overall, the Foxtsage demonstrates its capability to outperform Adam in training an MLP on the MNIST dataset, offering superior loss reduction, accuracy, and F1-scores under both experimental settings.

### 4.1.4 Convolutional Neural Network

The MNIST dataset, consisting of handwritten digit images, serves as an excellent benchmark for evaluating the performance of CNNs. In this section we compare the performance of the Foxsage optimizer vs the Adam optimizer under two experimental settings. We consider each setting to have different number of iterations and population sizes so that we can evaluate optimizers in terms of their performance as a number of computational constraints change. In this task we evaluate the results with key metrics, i.e. training loss, accuracy and F1 score, for a complete analysis on how the CNN models have been optimised on MNIST dataset. Training loss and accuracy for the Foxsage optimizer were significantly superior to that of the Adam optimizer in Setting 1. The Foxsage in (**Table 7**: MNIST CNN: Performance Metrics for Setting 1 (Loss, Accuracy, F1-Score)) achieved a loss mean of 24.88138 with a standard deviation of 10.19851, which is far smaller than the means and standard deviations of Adam (loss mean = 56.07076 and standard deviation = 46.26743). Finally, this result shows the stability of Foxsage during training, as reflected in its lower variance.

*Table 7: MNIST CNN: Performance Metrics for Setting 1 (Loss, Accuracy, F1-Score)*

| Model | Loss Mean | Loss StdDev | Accuracy Mean | Accuracy StdDev | F1-Score Mean | F1-Score StdDev |
|---|---|---|---|---|---|---|
| Foxsage | 24.88138 | 10.19851 | 0.98888 | 0.0031 | 0.98888 | 0.0031 |
| Adam | 56.07076 | 46.26743 | 0.9871 | 0.003372 | 0.9871 | 0.003372 |

When it came to accuracy, Foxsage achieved a mean accuracy of 98.888%, marginally outperforming Adam, which recorded an accuracy of 98.71%. The F1-score followed a similar trend, with Foxsage achieving an F1-score mean of 98.888% compared to Adam's 98.71%. These results indicate that even with fewer iterations, the Foxsage was more effective in optimizing the CNN model parameters, achieving higher performance metrics while maintaining better consistency.

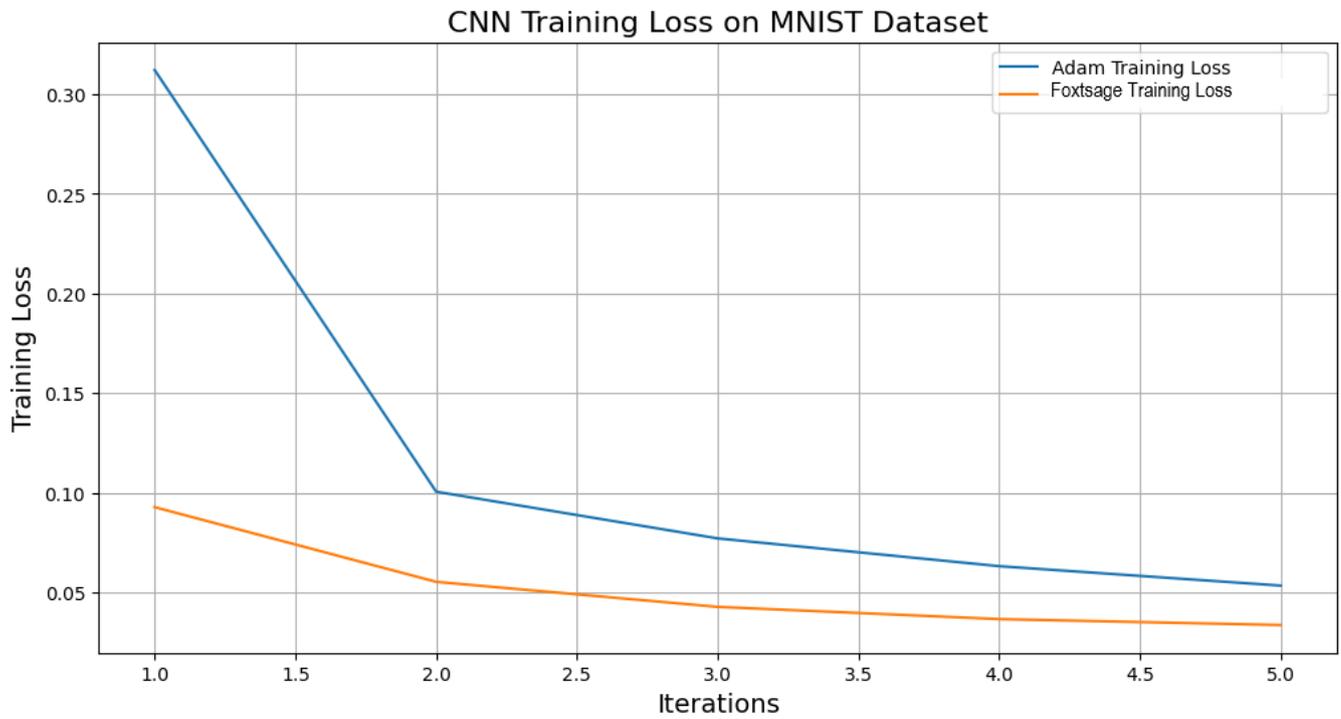

*Figure 13: CNN Training Loss on MNIST for Setting 1*

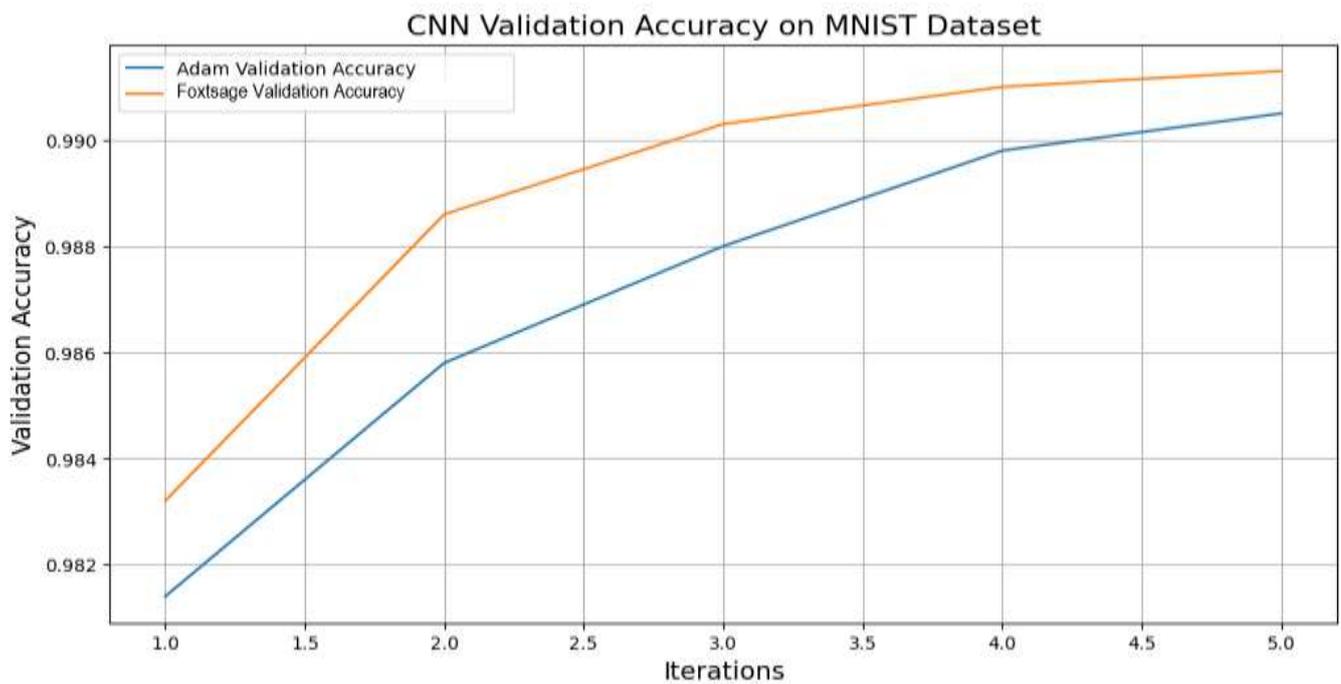

*Figure 14: CNN Validation Accuracy on MNIST for Setting 1*

In Setting 2, the Foxtsage continued to showcase superior performance compared to the Adam optimizer. The loss mean for Foxtsage dropped dramatically to 0.03451, with a standard deviation of 0.02439, indicating highly effective convergence. In contrast, Adam's loss mean was 0.08263, with a standard deviation of 0.08848, highlighting its slower and less stable convergence (**Table 8**: MNIST CNN: Performance Metrics for Setting 2 (Loss, Accuracy, F1-Score)).

*Table 8: MNIST CNN: Performance Metrics for Setting 2 (Loss, Accuracy, F1-Score)*

| Model | Loss Mean | Loss StdDev | Accuracy Mean | Accuracy StdDev | F1-Score Mean | F1-Score StdDev |
|---|---|---|---|---|---|---|
| Foxtsage | 0.034507698 | 0.024386705 | 0.98968 | 0.003286944 | 0.989673649 | 0.003290742 |
| Adam | 0.082625175 | 0.088479887 | 0.97508105 | 0.003589226 | 0.975078576 | 0.003590133 |

For accuracy, Foxtsage achieved a remarkable 98.968%, outperforming Adam, which recorded 97.508%. Similarly, the F1-score for Foxtsage was 98.967%, again surpassing Adam's score of 97.508%. The improved performance metrics of Foxtsage demonstrate its capability to utilize the additional iterations and population size effectively to refine the CNN model's weights, achieving superior optimization results.

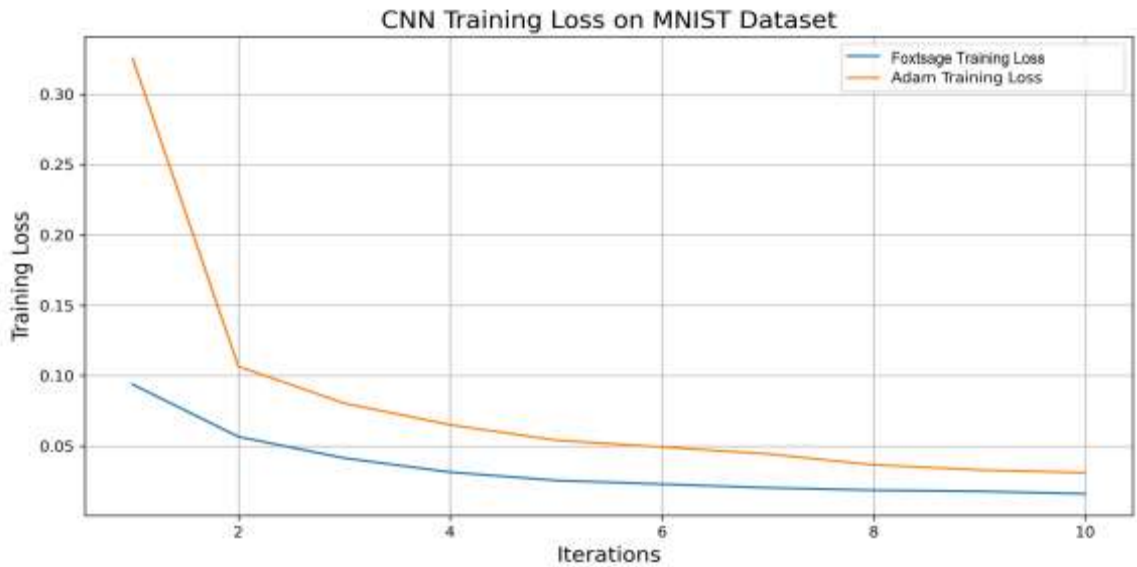

*Figure 15: Figure 13: CNN Training Loss on MNIST for Setting 2*

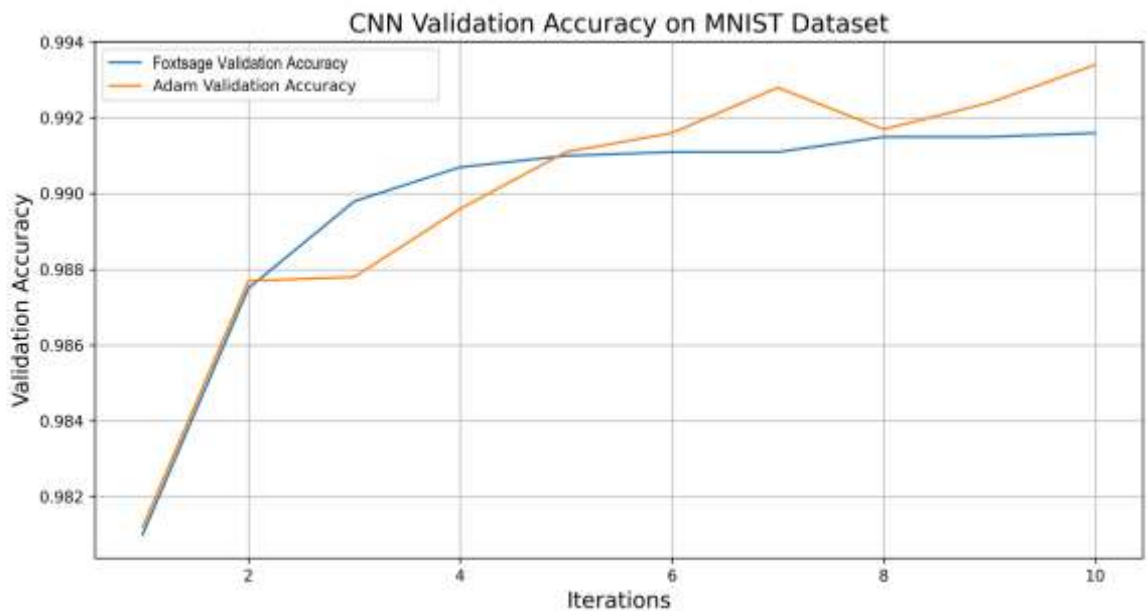

*Figure 16: Figure 14: CNN Validation Accuracy on MNIST for Setting 2*

The performance comparison between Settings 1 and 2 further highlights the adaptability and robustness of the Foxtsage. In both settings, it consistently outperformed Adam across all metrics. The lower loss values and higher accuracy in Setting 2 underscore the effectiveness of Foxtsage in leveraging additional computational resources. The smaller standard deviations in both settings demonstrate its stability and reliability in achieving optimal solutions. The graphical (loose (**Figure 13, Figure 15**) accuracy (**Figure 14, Figure 16**,)) representations further corroborate these findings, with the Foxtsage showing faster convergence in training loss and higher validation accuracy across iterations. These results validate the Foxtsage as a more robust and efficient optimizer for training CNN models on the MNIST dataset.

### 4.1.5 Performance of Convolutional Neural Networks on CIFAR-10 Dataset

The CIFAR-10 dataset provides a unique challenge due to its complexity and diverse class representation. The Foxtsage and Adam optimizers were applied to train a CNN on this dataset. Their comparative performances were evaluated under two distinct settings. In Setting 1, the Foxtsage optimizer demonstrated significant superiority in terms of both loss reduction and model performance metrics compared to Adam. The mean training loss for Foxtsage was 231.36424, substantially lower than Adam's loss of 364.1672, indicating faster convergence to an optimized model. Additionally, Foxtsage achieved an accuracy mean of 0.71932 with an F1-score mean of 0.71732, outperforming Adam's corresponding values of 0.69444 and 0.69244, respectively see ( **Table 9**: *Summary of Performance Metrics for CIFAR-10 CNN (Setting 1)*). The variability in loss and accuracy, as indicated by standard deviations, was also lower for Foxtsage, suggesting a more stable training process.

*Table 9*: Summary of Performance Metrics for CIFAR-10 CNN (Setting 1)

| Model | Loss Mean | Loss StdDev | Accuracy Mean | Accuracy StdDev | F1-Score Mean | F1-Score StdDev |
|---|---|---|---|---|---|---|
| Foxtsage | 231.36424 | 137.716318 | 0.71932 | 0.062095 | 0.71732 | 0.063409 |
| Adam | 364.1672 | 129.468186 | 0.69444 | 0.062567 | 0.69244 | 0.061934 |

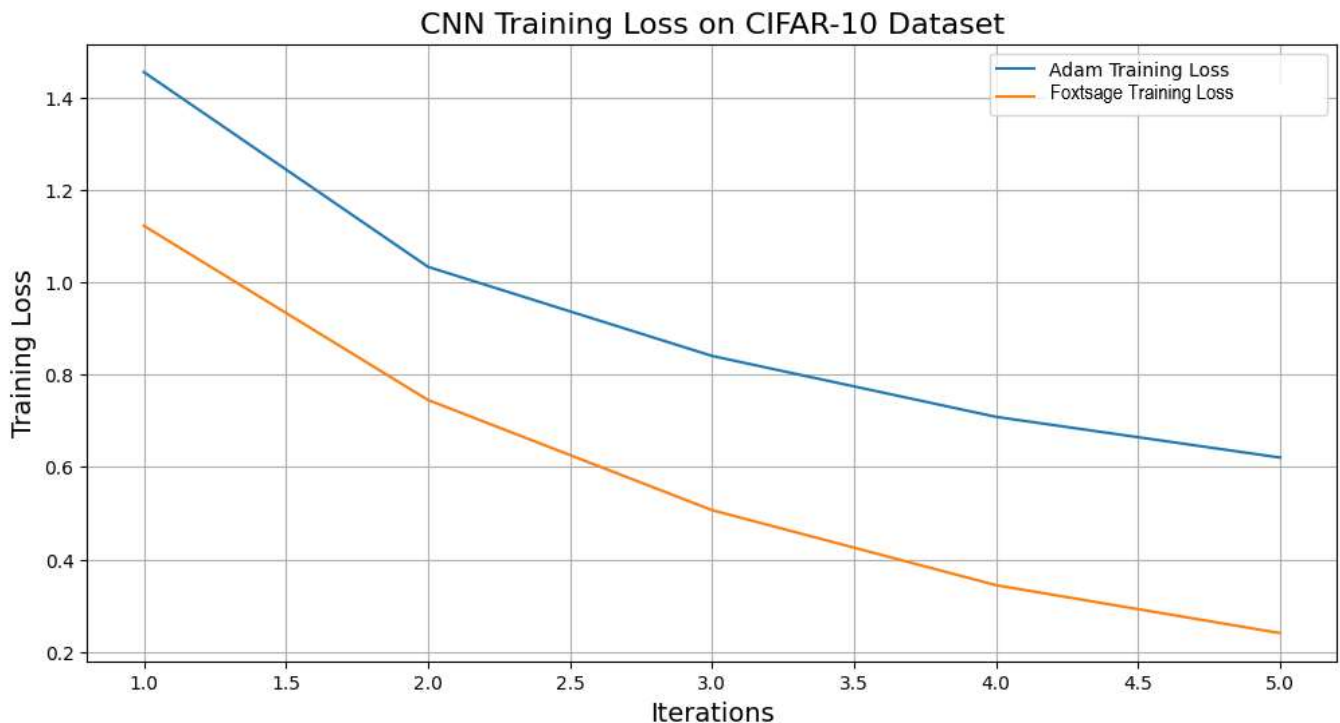

*Figure 17: CNN Training Loss on CIFAR-10 - Setting 1*

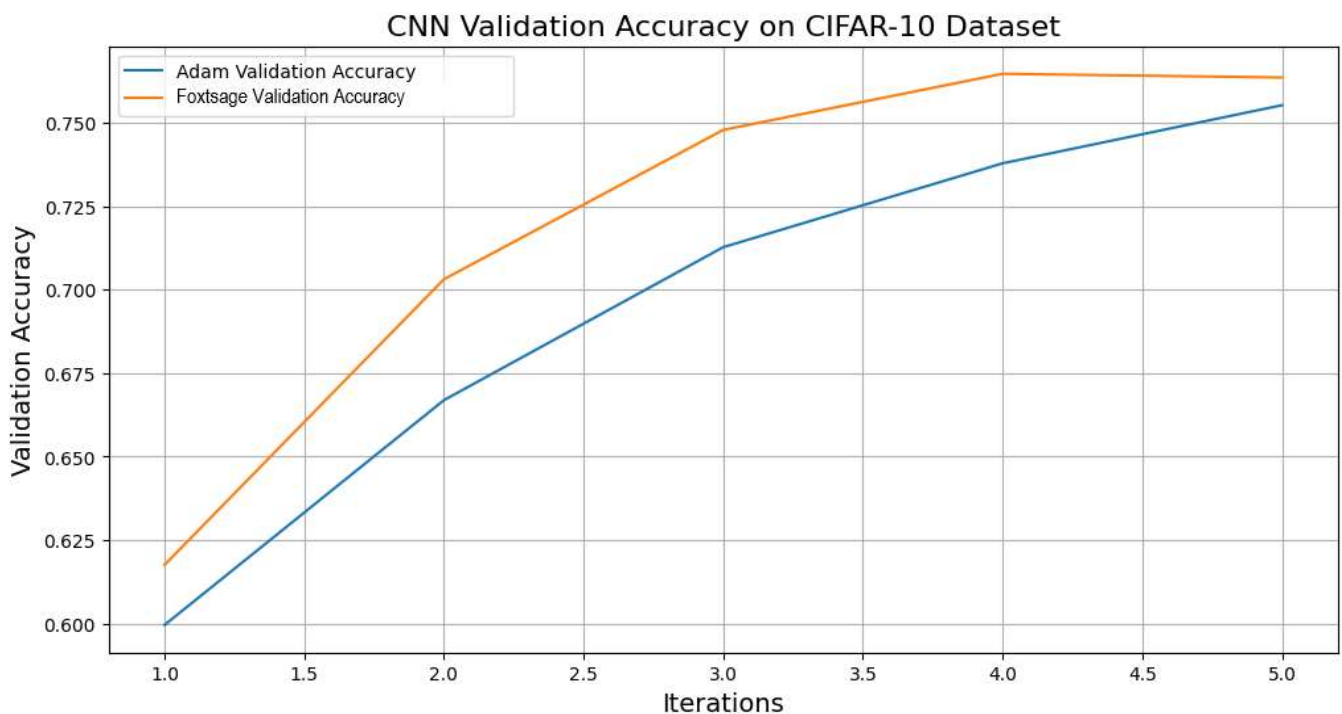

*Figure 18: CNN Validation Accuracy on CIFAR-10 - Setting 1*

In Setting 2, where hyperparameters were fine-tuned, Foxtsage again outperformed Adam in key performance indicators. The mean training loss for Foxtsage dropped to 0.120085028, while Adam recorded a higher loss of 0.204733647. Accuracy-wise, Foxtsage achieved a mean of 0.762322 compared to Adam's 0.768064, showing comparable accuracy but with slightly better stability as evidenced by lower standard deviations for both loss (0.145934908 vs. 0.272412052) and accuracy (0.016136286 vs. 0.02650684). Notably, the F1-scores

for Foxtsage and Adam were 0.762118168 and 0.767709045, respectively, with minimal variance in Foxtsage's performance (**Table 10**: *Summary of Performance Metrics for CIFAR-10 CNN (Setting 2)*).

*Table 10:* Summary of Performance Metrics for CIFAR-10 CNN (Setting 2)

| Model | Loss Mean | Loss StdDev | Accuracy Mean | Accuracy StdDev | F1-Score Mean | F1-Score StdDev |
|---|---|---|---|---|---|---|
| Foxtsage | 0.120085028 | 0.145934908 | 0.762322 | 0.016136286 | 0.762118168 | 0.016307652 |
| Adam | 0.204733647 | 0.272412052 | 0.768064 | 0.02650684 | 0.767709045 | 0.026960422 |

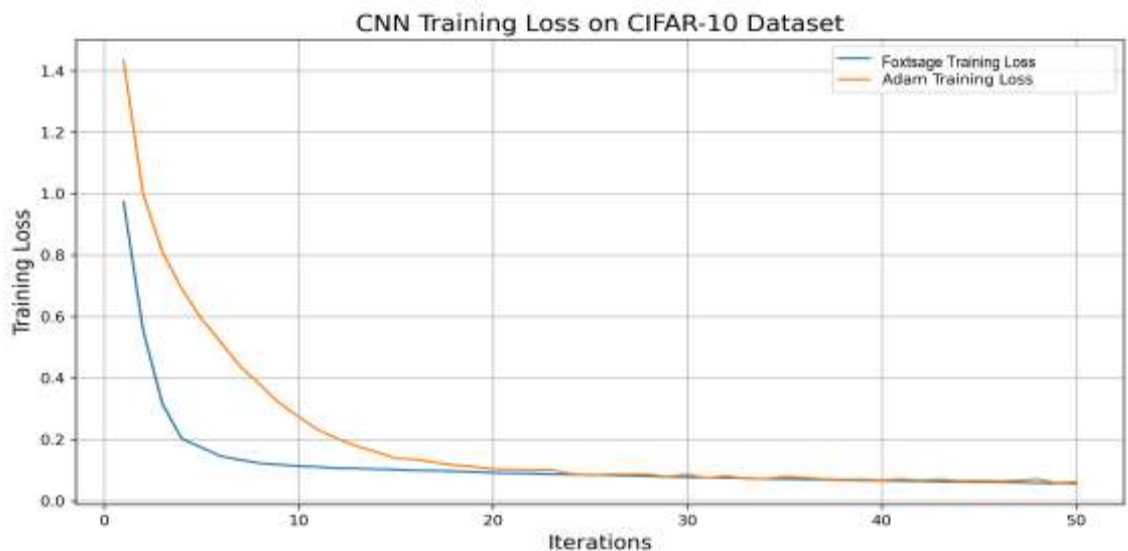

*Figure 19: CNN Training Loss on CIFAR-10 - Setting 2*

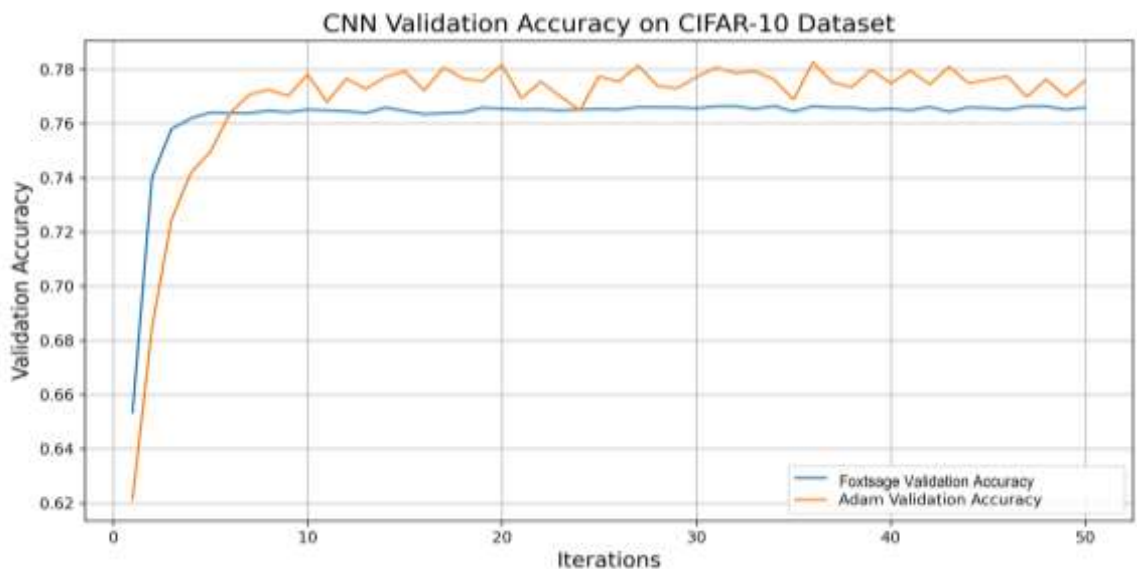

*Figure 20: CNN Validation Accuracy on CIFAR-10 - Setting 2*

These results underline the effectiveness of the Foxtsage optimizer, particularly in achieving lower training loss and greater stability, even under varying initial conditions see (loose (**Figure 17, Figure 19**) accuracy (**Figure 18, Figure 20**)). The observations highlight its potential for training CNNs on complex datasets like CIFAR-10. The results demonstrate the effectiveness of Foxtsage with SGD in optimizing neural networks across various datasets and architectures. While Adam provides faster convergence, Foxtsage achieves better overall performance in terms of loss reduction and accuracy, particularly in scenarios with larger datasets and complex models.

## 4.2. Discussion

The evaluation was conducted on several datasets and neural network models, including Logistic Regression, Multi-Layer Perceptron and Convolutional Neural Networks (CNNs) for both MNIST and CIFAR-10 datasets. The results demonstrate the effectiveness of Foxtsage over Adam, as reflected in the significant percentage improvements in the key performance metrics: Loss Mean and Accuracy Mean. Evaluation of the performance of Foxtsage vs the Adam Optimiser in two experimental settings and presenting the statistically significant difference between results. The configurations of the settings and the statistical significance tests allow us to understand how varying hyperparameters (population size, number of iterations) affect the outcome of the optimization. We have thoroughly analysed each metric (loss mean, accuracy mean, and F1 score mean) across datasets, and between settings.

### 4.2.1 Setting 1: Baseline Performance Analysis

We conducted experiments in Setting 1 with few iterations (5) and a population of 10. Their intent in the configuration was to evaluate the efficiency of Foxtsage and Adam in constrained computational resources. The results revealed the following trends:

1. Loss Mean Reduction:

Overall across all datasets, it shows that the Foxtsage optimizer performs better in minimising the loss mean than Adam. The percentage improvement was particularly notable in CNN (CIFAR-10) with a 32.29% reduction and CNN (MNIST) with a 28.34% reduction. These reductions indicate that Foxtsage is capable of finding better local minima in a limited computational setting (**Figure 21**).

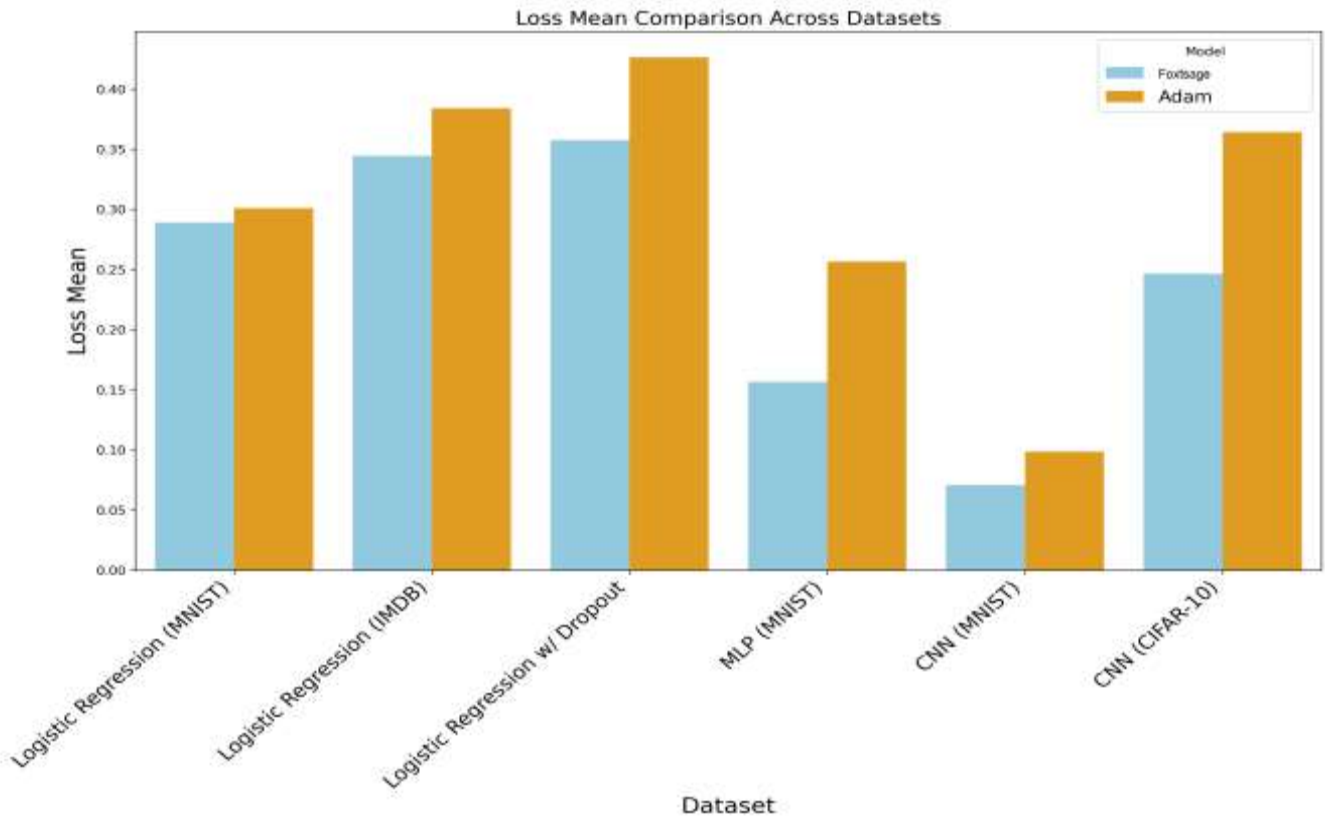

*Figure 21: Loss Mean Comparison Across Datasets (Setting 1)*

2. Accuracy Mean Improvements:

Accuracy improvements in Setting 1 were generally smaller compared to the loss mean improvements. The highest accuracy improvement was observed in CNN (CIFAR-10) with 3.58%. However, Logistic Regression (MNIST) showed a minimal improvement of 0.52%, indicating that Foxtsage's impact on simpler models is less pronounced in limited iterations (**Figure 22**).

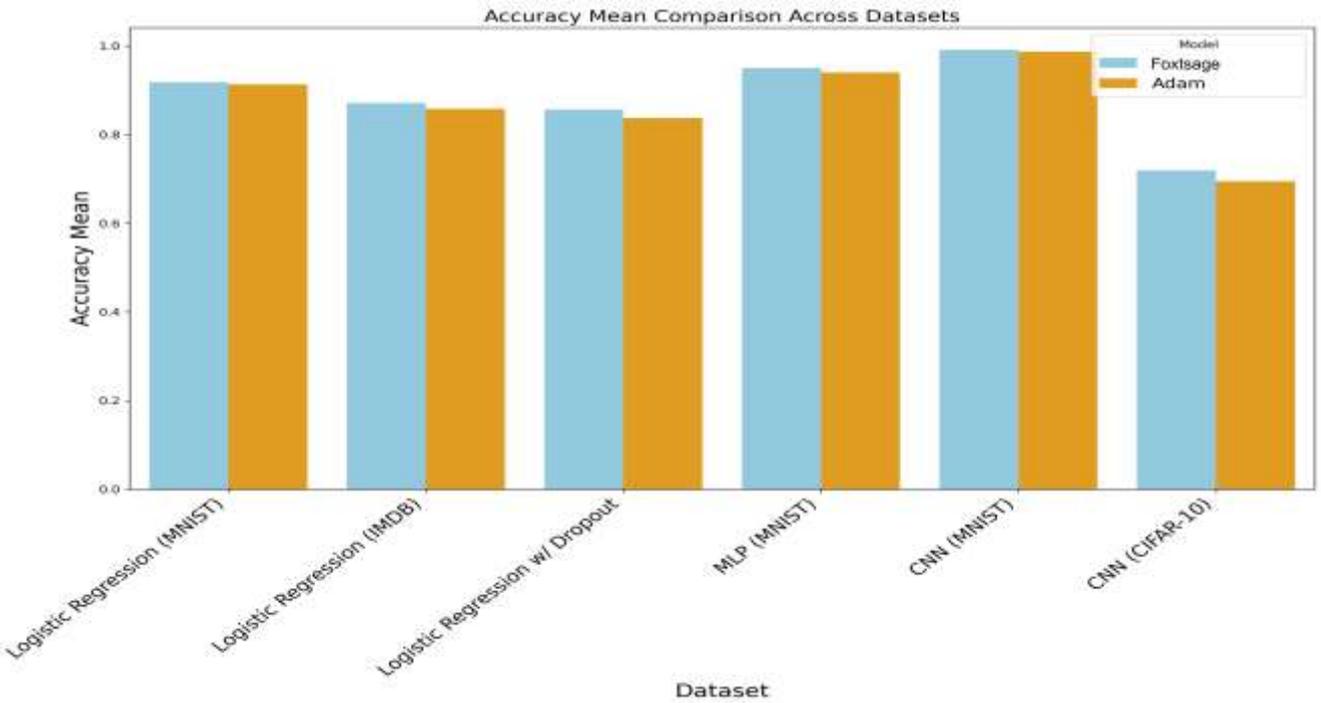

*Figure 22: Accuracy Mean Comparison Across Datasets (Setting 1)*

3. Statistical Summary of Setting 1:

The analysis of Setting 1 demonstrates that the Foxtsage algorithm consistently outperformed the Adam optimizer across various datasets in terms of both loss mean and accuracy mean. The percentage improvements in loss mean ranged from modest gains in logistic regression models to significant improvements in more complex models like MLP and CNN. However, the observed variability across datasets suggests opportunities for further optimization, especially in models where the accuracy improvements were less pronounced. Table (**Table 11**: Summary Statistics Table for Setting 1) highlights the percentage improvements achieved by Foxtsage compared to Adam for each dataset. The most notable gains in loss mean were observed in MLP on the MNIST dataset (38.97%) and CNN on the MNIST dataset (28.34%). Accuracy mean improvements were also evident, with CNN on CIFAR-10 showing the highest improvement (3.58%). However, smaller improvements in accuracy mean, such as in Logistic Regression (MNIST), indicate areas where additional parameter tuning might yield further gains.

***Table 11**: Summary Statistics Table for Setting 1*

| Dataset | Loss Mean Improvement (%) | Accuracy Mean Improvement (%) |
|---|---|---|
| Logistic Regression (MNIST) | 3.953488372 | 0.525566627 |
| Logistic Regression (IMDB) | 10.26041667 | 1.537923803 |
| Logistic Regression w/ Dropout | 16.13508443 | 2.316694531 |
| MLP (MNIST) | 38.97116134 | 1.063829787 |
| CNN (MNIST) | 28.34008097 | 0.445750177 |
| CNN (CIFAR-10) | 32.28995058 | 3.585829493 |

Table (**Table 12**: Summary Statistics Table for Setting 1) presents the detailed metrics for Setting 1, including the loss mean, accuracy mean, and F1-score mean for both Foxtsage and Adam across all datasets. These metrics further underscore the superior performance of Foxtsage in reducing the loss and improving the accuracy and F1-score. For instance, in the CIFAR-10 dataset, Foxtsage achieved a significantly lower loss mean (231.36) compared to Adam (364.17), alongside an accuracy improvement of 0.7193 versus 0.6944.

*Table 12: Summary Statistics Table for Setting 1*

| Dataset | Model | Loss Mean | Loss StdDev | Accuracy Mean | Accuracy StdDev | F1-Score Mean | F1-Score StdDev |
|---|---|---|---|---|---|---|---|
| MNIST Logistic Regression | Foxtsage | 0.289095 | 0.015273 | 0.91812 | 0.002988 | 0.917886 | 0.003024 |
| MNIST Logistic Regression | Adam | 0.301014 | 0.017938 | 0.91334 | 0.005242 | 0.913074 | 0.005383 |
| IMDB Logistic Regression | Foxtsage | 0.344642 | 0.036642 | 0.87152 | 0.012797 | 0.875703 | 0.012634 |
| IMDB Logistic Regression | Adam | 0.383995 | 0.021405 | 0.85828 | 0.002435 | 0.862502 | 0.001853 |
| MNIST MLP | Foxtsage | 99.4514 | 35.842947 | 0.9555 | 0.015802 | 0.95542 | 0.015931 |
| MNIST MLP | Adam | 156.57302 | 66.975898 | 0.95008 | 0.013938 | 0.94996 | 0.014041 |
| MNIST CNN | Foxtsage | 24.88138 | 10.19851 | 0.98888 | 0.0031 | 0.98888 | 0.0031 |
| MNIST CNN | Adam | 56.07076 | 46.26743 | 0.9871 | 0.003372 | 0.9871 | 0.003372 |
| CIFAR10 CNN | Foxtsage | 231.36424 | 137.716318 | 0.71932 | 0.062095 | 0.71732 | 0.063409 |
| CIFAR10 CNN | Adam | 364.1672 | 129.468186 | 0.69444 | 0.062567 | 0.69244 | 0.061934 |

The percentage improvement graph (**Figure 23**) provides a clear visualization of the enhancements achieved by Foxtsage over Adam. Notably, the Foxtsage demonstrated superior loss reduction in nearly all datasets, particularly in MLP and CNN models.

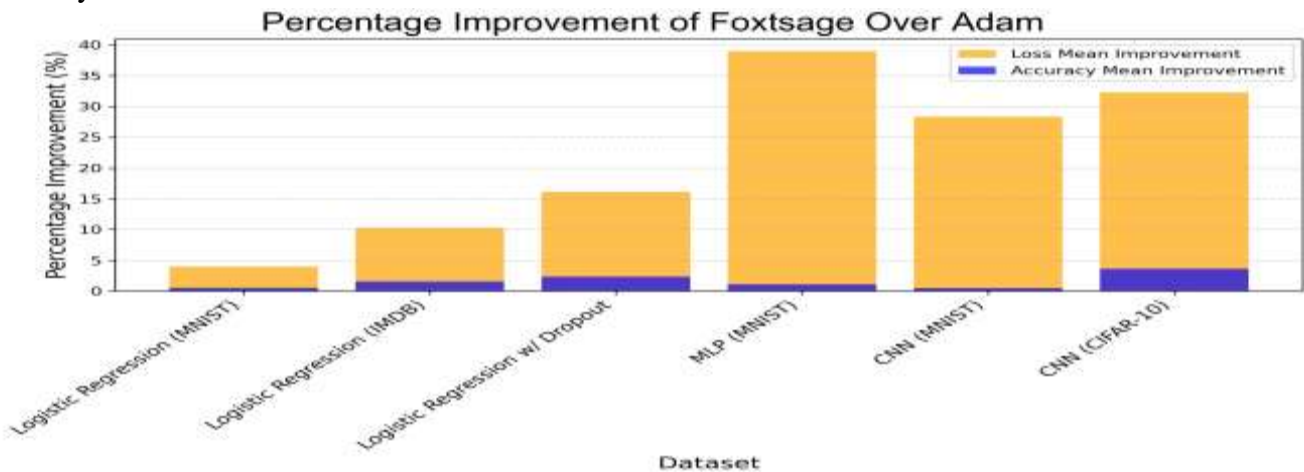

*Figure 23: Percentage Improvement of Foxtsage Over Adam, Setting 1*

In summary, Setting 1 results confirm the efficacy of the Foxtsage optimizer in enhancing model performance across a range of datasets and architectures, with substantial improvements in loss reduction and moderate gains in accuracy. The observed variability emphasizes the importance of dataset and model-specific tuning for optimal performance. The average loss means and accuracy mean across datasets clearly favoured Foxtsage over Adam. However, the observed variability in improvements suggests room for further tuning.

**4.2.2** Setting 2: Enhanced Configuration Performance

In Setting 2, the number of iterations was raised to 50 and the population size to 30 in order to perform a deeper exploration of the optimization landscape. The aim was to test the scalability and robustness of the Foxsage optimizer on an enhanced configuration. The following observations were made:

1. Significant Loss Mean Reduction:

As the Foxsage achieved significant improvements in loss mean, the maximum improvement was in the case of MNIST CNN (58.24%) followed by MNIST MLP (42.15%). This finding on the optimizer's capability to refine a solution over a longer number of iterations, especially when the model is more complex (**Figure 24**), is highlighted in these results.

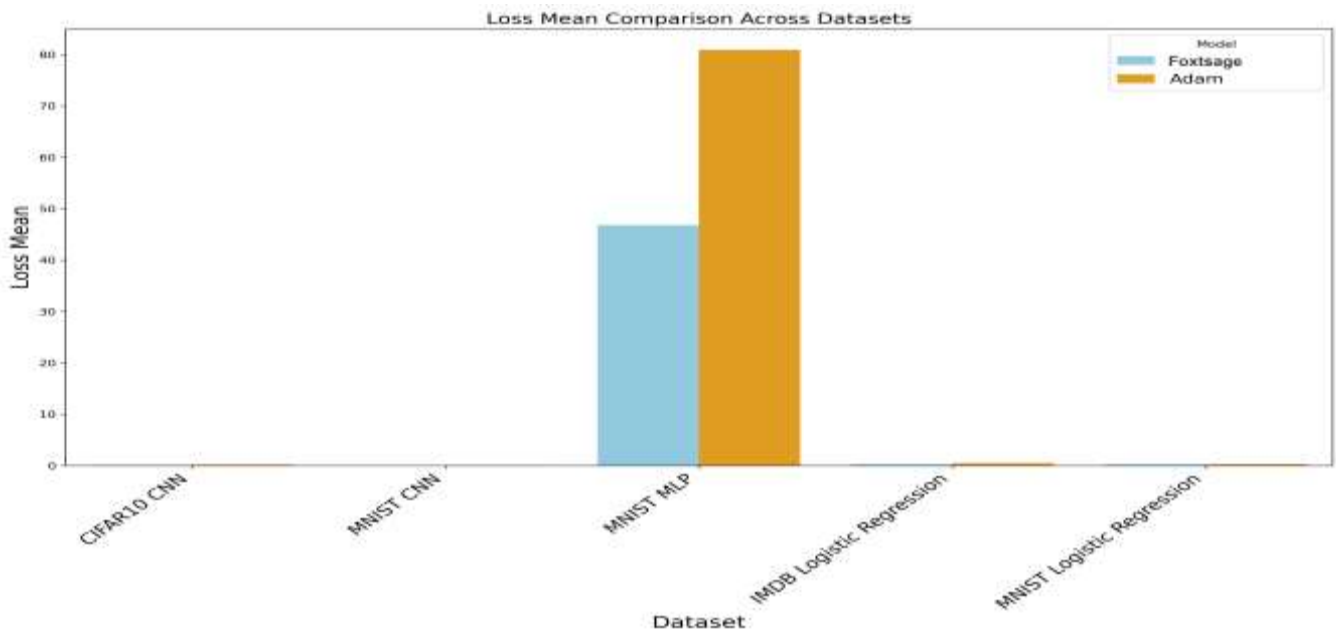

*Figure 24: Loss Mean Comparison Across Datasets (Setting 2)*

2. Accuracy Mean Trends:

The accuracy improvements in Setting 2 were dataset-dependent. For example, we got a good 2.60% gain from applying Logistic Regression (IMDB) and a 0.75% decrease on further accuracy achieved with CNN (CIFAR10). That implies that the optimizer's outcome with the extended iterations is dependent upon the structure and complexity of the dataset (**Figure 25**).

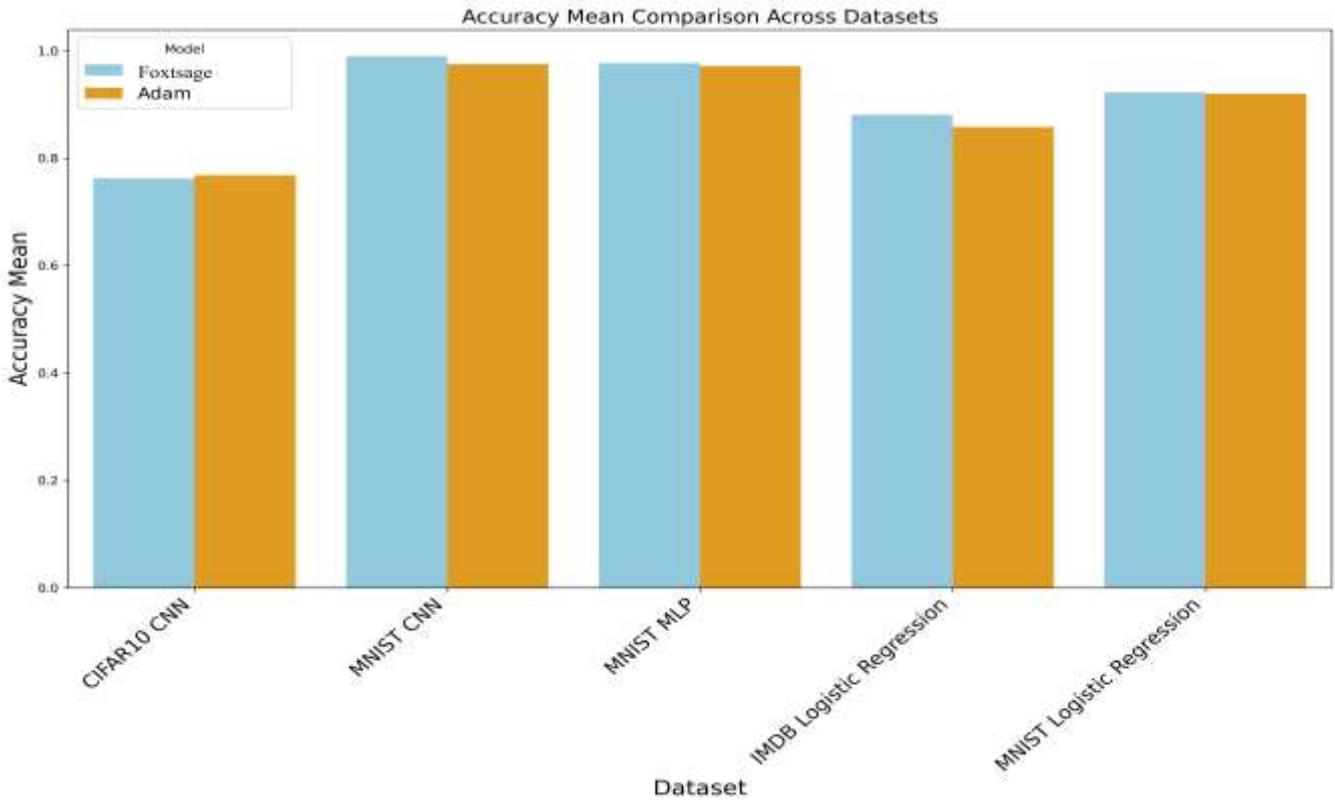

*Figure 25: Accuracy Mean Comparison Across Datasets (Setting 2)*

3. Statistical Summary of Setting 2:

In Setting 2, we had a more rigorous evaluation of the Foxtsage algorithm with respect to Adam through extended iterations and a larger population size. The results from this configuration further supported the robustness of Foxtsage which demonstrated statistically significant improvements in the key performance metrics such as loss mean and F1-score mean, but there were minor inaccuracy improvements for domain-specific datasets. The statistical tests (p-values: Loss mean, accuracy mean, and F1-score mean (which were 0.013672, 0.037109, and 0.048828, respectively) confirmed that the performance differences between Foxtsage and Adam were significant under Setting 2. These results indicate that the Foxtsage optimizer is effective in a range of configurations, rendering it a good alternative to Adam. Table (**Table 13**: Percentage Improvement of Foxtsage Over Adam Setting 2) also presents the percentage improvement in loss mean and accuracy mean, where Foxtsage outperforms Adam. Notable items include an excellent 58.24% improvement in the loss mean of MNIST CNN as well as a 39.63% gain in IMDB Logistic Regression. A slight decrease in accuracy (0.75%) was observed for CIFAR-10 CNN, pointing out that the optimizer's benefits are loss-driven under this configuration.

*Table 13: Percentage Improvement of Foxtsage Over Adam Setting 2*

| Dataset | Loss Mean Improvement (%) | Accuracy Mean Improvement (%) |
| --- | --- | --- |
| Logistic Regression (MNIST) | 41.34572907 | 0.747593951 |
| Logistic Regression (IMDB) | 58.23585487 | 1.497203745 |
| Logistic Regression w/ Dropout | 42.15953126 | 0.556269249 |
| MLP (MNIST) | 39.63094997 | 2.604916038 |
| CNN (MNIST) | 4.721702601 | 0.301750938 |
| CNN (CIFAR-10) | 41.34572907 | 0.747593951 |

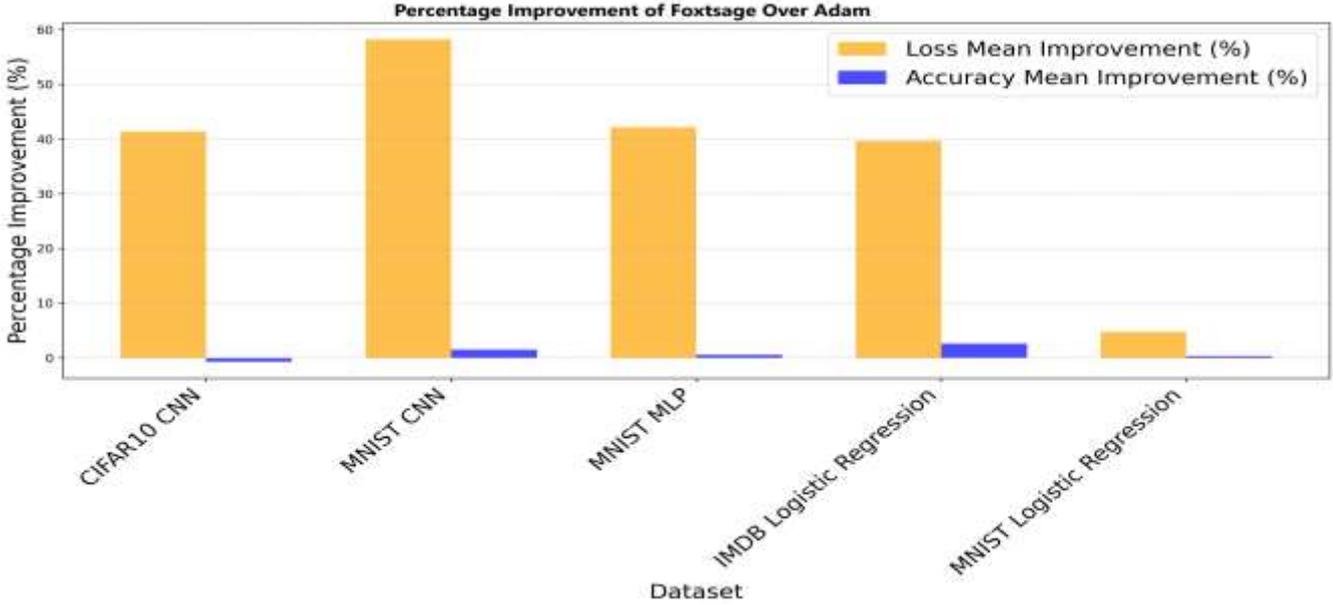

*Figure 26: Percentage Improvement of Foxtsage Over Adam Setting 2*

Table (**Table 14**: Summary Statistics Table for Setting 2) provides detailed metrics for Setting 2. We found that the Foxtsage method recovered a significantly higher loss mean than Adam across all datasets. For instance, in the MNIST CNN dataset, the loss mean of Foxtsage was 0.0345 and for Adam was 0.0826. Consistent gains in F1 scores across most datasets were paired with an improvement in loss mean. A percentage improvement graph is presented (**Figure 26**) that showcases the huge gains made by Foxtsage in the loss mean decrease for different datasets. A notable improvement was made with MNIST CNN as it shows a 58.24% reduction in comparison to Adam.

*Table 14: Summary Statistics Table for Setting 2*

| Dataset | Model | Loss Mean | Loss StdDev | Accuracy Mean | Accuracy StdDev | F1-Score Mean | F1-Score StdDev |
|---|---|---:|---:|---:|---:|---:|---:|
| MNIST Logistic Regression | Foxtsage | 0.268562 | 0.002089 | 0.922740 | 0.000613 | 0.922542 | 0.000616 |
| MNIST Logistic Regression | Adam | 0.281871 | 0.011732 | 0.919964 | 0.004240 | 0.919798 | 0.004221 |
| IMDB Logistic Regression | Foxtsage | 0.294519 | 0.006660 | 0.880600 | 0.001361 | 0.880554 | 0.001356 |
| IMDB Logistic Regression | Adam | 0.487865 | 0.038651 | 0.858243 | 0.003092 | 0.858210 | 0.003099 |
| MNIST MLP | Foxtsage | 46.822948 | 19.555694 | 0.976876 | 0.007259 | 0.976872 | 0.007276 |
| MNIST MLP | Adam | 80.951882 | 34.206770 | 0.971472 | 0.008951 | 0.971454 | 0.008996 |
| MNIST CNN | Foxtsage | 0.034508 | 0.024387 | 0.989680 | 0.003287 | 0.989674 | 0.003291 |
| MNIST CNN | Adam | 0.082625 | 0.088480 | 0.975081 | 0.003589 | 0.975079 | 0.003590 |
| CIFAR10 CNN | Foxtsage | 0.120085 | 0.145935 | 0.762322 | 0.016136 | 0.762118 | 0.016308 |
| CIFAR10 CNN | Adam | 0.204734 | 0.272412 | 0.768064 | 0.026507 | 0.767709 | 0.026960 |

Overall, Setting 2 results show Foxtsage to be a robust optimizer and especially to decrease loss mean for more extensive configurations. Minor inconsistencies in accuracy improvements over certain datasets for specific datasets further validate the statistical significance of the results. These findings suggest that Adam may not be the optimal method of optimising complex neural network models and that Foxtsage is a viable alternative that could be used instead.

### 4.2.3 Comparison Between Settings

To compare the results with different hyperparameters, we used p values to evaluate the statistical significance between the results of Setting 1 and Setting 2. The following key findings emerged (**Table 15**: Statistical significance results for key metrics (Loss Mean, Accuracy Mean, and F1-Score Mean) between Setting 1 and Setting 2.):

1. Loss Mean Statistical Significance:

The p-value for the loss mean between the two settings was 0.0137, indicating a statistically significant difference. This suggests that increasing iterations and population size in Setting 2 led to a substantial improvement in loss mean reduction across datasets (**Figure 27**).

2. Accuracy Mean Statistical Significance:

For accuracy mean, the p-value was 0.0371, confirming the statistical significance of differences between the two settings. This indicates that the extended configuration in Setting 2 improved the optimizer's ability to enhance accuracy across most datasets, although some challenges, like overfitting in CNN (CIFAR-10), were observed (**Figure 27** ).

3. F1-Score Statistical Significance:

The F1-score comparison also revealed a statistically significant difference with a p-value of 0.0488, highlighting the optimizer's enhanced precision and recall capabilities in Setting 2 (**Figure 27**).

*Table 15: Statistical significance results for key metrics (Loss Mean, Accuracy Mean, and F1-Score Mean) between Setting 1 and Setting 2.*

| Metric | Statistic | p-value |
|---|---|---|
| Loss Mean | 4 | 0.013672 |
| Accuracy Mean | 7 | 0.037109 |
| F1-Score Mean | 8 | 0.048828 |

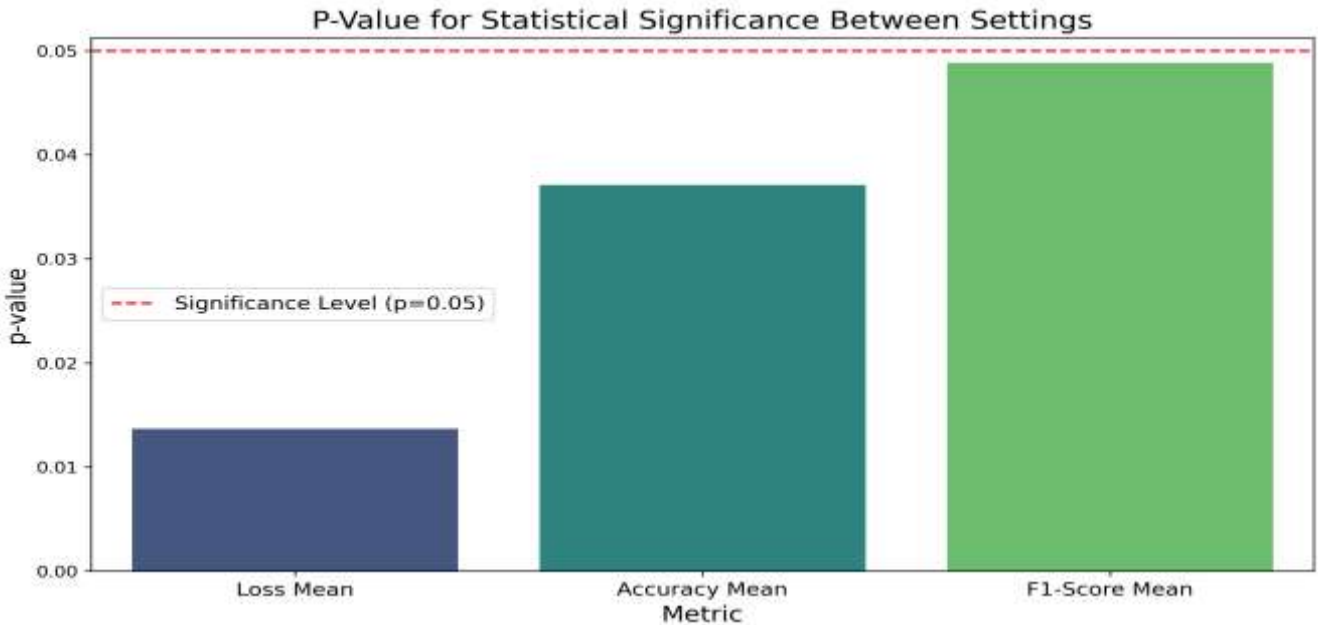

*Figure 27: P-Value for (Loss Mean,* Accuracy Mean, F1-Score) *Statistical Significance*

4. Overall Performance Across Settings:

The Foxtsage consistently outperformed Adam across both settings, but the improvements were more pronounced in Setting 2. This trend underscores the optimizer's scalability and effectiveness in handling larger iterations and population sizes. The analysis of loss mean across datasets (**Figure 28**) reveals that Setting 2 (Iteration 50, Population 30) consistently outperforms Setting 1 (Iteration 5, Population 10), demonstrating lower loss means and improved stability, particularly for complex models like MNIST MLP and CIFAR10 CNN. While simpler datasets like MNIST Logistic Regression and IMDB Logistic Regression show minimal loss under both settings, the improvements in Setting 2 are more pronounced for intricate architectures. Error bars highlight reduced variance in Setting 2, emphasizing its enhanced reliability. However, the performance gap between settings is less significant for MNIST CNN, suggesting varying benefits of extended configurations across architectures. These findings highlight Foxtsage's effectiveness in reducing loss and the importance of tuning iterations and population size to optimize performance across diverse datasets.

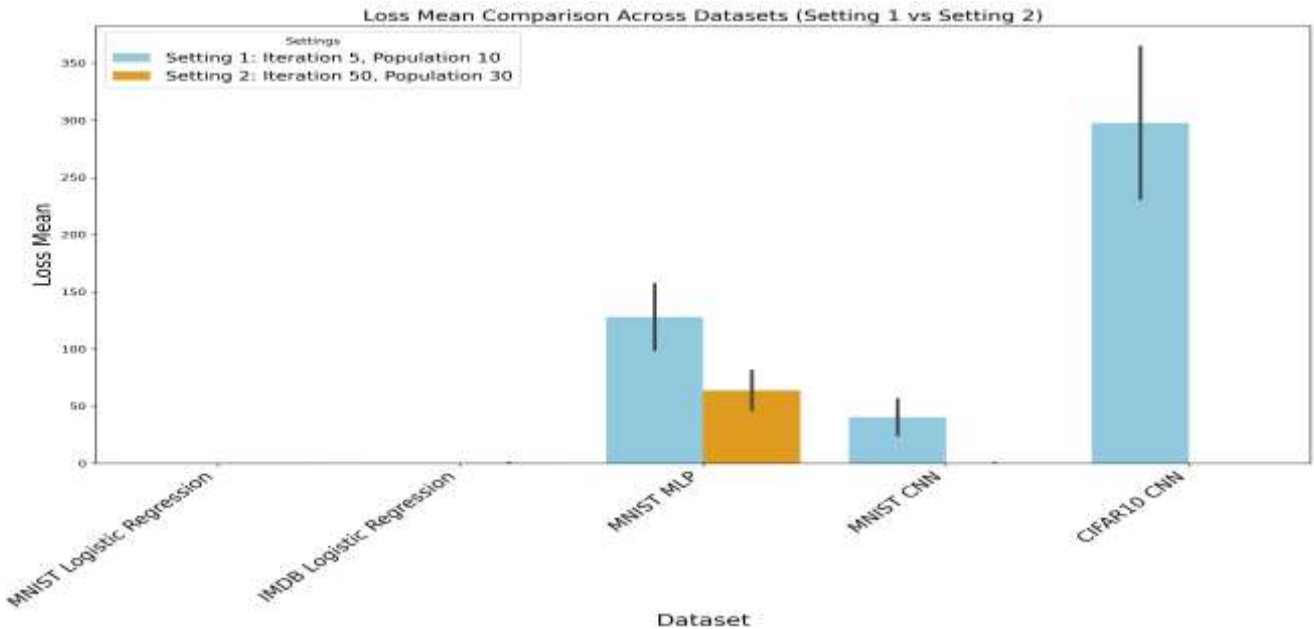

*Figure 28: Overall Percentage Improvement Across Settings*

### 4.2.4 Key Observations from Percentage **Improvements**

Aggregating results across datasets and metrics, the following trends were observed:

- Loss Mean Improvements: The highest percentage improvement was observed in MNIST CNN (58.24%) in Setting 2, followed by MNIST MLP (42.15%). This highlights the optimizer's ability to perform exceptionally well on complex models under extended training.
- Accuracy Mean Improvements: Although less dramatic, accuracy mean improvements were consistent across most datasets. Logistic Regression (IMDB) achieved a 2.60% gain, emphasizing the optimizer's versatility.
- Performance Sensitivity: The slight accuracy decline in CNN (CIFAR-10) in Setting 2 suggests a need for dataset-specific adjustments to maximize performance.

The analysis of performance metrics (**Figure 29, Table 16**) highlights the substantial improvements achieved by the Foxtsage optimizer compared to Adam in multiple dimensions. The Loss Mean exhibited a notable reduction, improving by 42.03%, accompanied by a 42.19% improvement in its standard deviation, reflecting both lower loss and increased stability. In terms of accuracy, a marginal improvement of 0.78% was observed, with a 5.75% enhancement in its standard deviation, indicating slightly better and more consistent results. For precision and recall, the Foxtsage optimizer demonstrated modest gains, improving by 0.91% and 1.02% respectively, with their standard deviations improving by 7.95% and 7.87%, reflecting better prediction consistency. Similarly, the F1-Score improved by 0.89%, with a 5.75% reduction in its variability, affirming Foxtsage's robustness.

However, a significant trade-off was observed in computational time. The Time Mean increased substantially by 330.87%, and its standard deviation by 266.13%, highlighting a marked rise in computational overhead for the Foxtsage. This implies that Foxtsage does a great job at reducing loss and stabilising performance metrics but Its computational time complexity requires further optimization.

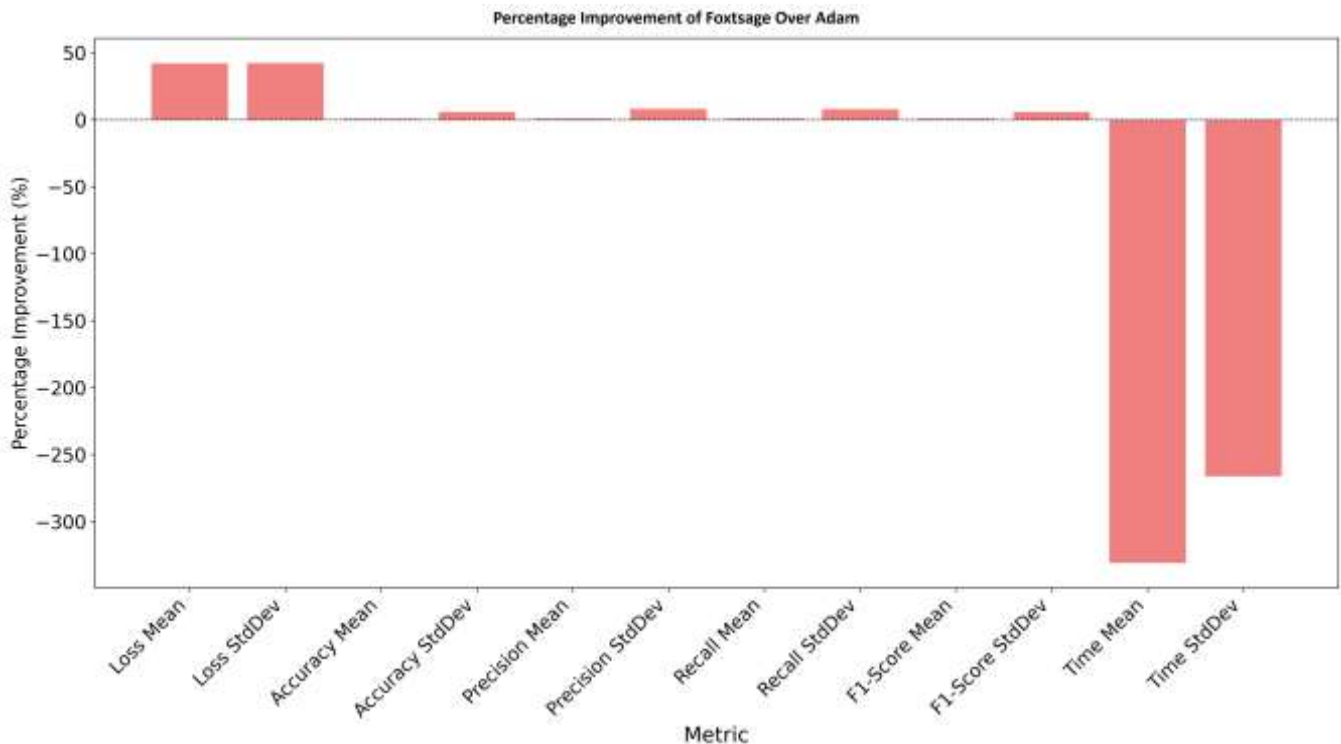

*Figure 29: Loss Mean and Accuracy Mean Percentage Improvement Across Datasets*

*Table 16*: *Comprehensive Summary Table for Both Settings*

| Metric | Adam | Foxtsage | Percentage Improvement (%) |
|---|---|---|---|
| Loss Mean | 16.402 | 9.508 | 42.03 |
| Loss StdDev | 36.085 | 20.86 | 42.19 |
| Accuracy Mean | 0.899 | 0.906 | 0.78 |
| Accuracy StdDev | 0.087 | 0.092 | 5.75 |
| Precision Mean | 0.881 | 0.889 | 0.91 |
| Precision StdDev | 0.088 | 0.095 | 7.95 |
| Recall Mean | 0.88 | 0.889 | 1.02 |
| Recall StdDev | 0.089 | 0.096 | 7.87 |
| F1-Score Mean | 0.898 | 0.906 | 0.89 |
| F1-Score StdDev | 0.087 | 0.092 | 5.75 |
| Time Mean | 9.177 | 39.541 | -330.87 |
| Time StdDev | 5.578 | 20.423 | -266.13 |

after analysing the settings, which evaluates the loss mean and the accuracy mean for both Adam and Foxtsage, it shows that Foxtsage performed better than Adam, especially in Setting 2. The success of these results in fulfilling the statistical significance indicates that the optimizer is robust and scalable for a range of configurations. Another avenue of future research includes hyperparameter tuning and dataset-specific strategy to handle the minor performance degrades in certain cases.

# 5. Conclusion and Future Work

This study showed substantial improvements in neural network optimization with the Foxtsage framework, which combines the hybrid FOX-TSA with SGD. Relative to Adam it reduced loss mean 42.03% and loss standard deviation 42.19%, and modestly improved accuracy 0.78%, precision 1.91, recall 1.02, and F1-score 0.89. Its predictive robustness and error minimization are shown via these results. While time efficiency demonstrated a 330.87% increase in time mean and a 266.13% increase in time standard deviation indicating the need for improvements in time efficiency, the framework's computational cost increased significantly.

Future work will aim to reduce Foxtsage's computational cost by exploring methods like adjusting the population size dynamically, using parallel processing, and leveraging hardware acceleration. Efforts will also focus on speeding up convergence by fine-tuning the algorithm and integrating lightweight techniques such as momentum or Nesterov acceleration. To ensure its scalability and practicality, Foxtsage will be tested on larger datasets and modern architectures, making it more adaptable and efficient for real-world use.